\pdfoutput=1  

\documentclass[12pt]{article}

\usepackage[T1]{fontenc}
\usepackage[utf8]{inputenc}

\usepackage[margin=1in]{geometry}

\usepackage{booktabs}
\usepackage{multirow}
\usepackage{amsmath,amssymb}
\usepackage{graphicx}
\usepackage{xcolor}
\usepackage{microtype}
\usepackage{tabularx}
\usepackage{array}
\usepackage{xurl}
\usepackage{authblk}
\usepackage{enumitem}
\usepackage{hyperref}
\hypersetup{colorlinks=true,linkcolor=blue,citecolor=blue,urlcolor=blue}

\definecolor{EavColor}{RGB}{0,100,160}
\definecolor{DtColor}{RGB}{34,139,34}
\definecolor{HelixColor}{RGB}{140,60,0}
\definecolor{RagColor}{RGB}{180,30,30}

\author[1]{Craig Atkinson}
\affil[1]{Verificate Pty Ltd, Sydney, Australia \\ \texttt{craig.atkinson@verificate.ai}}

\begin{document}

\title{Deterministic Decisions for High-Stakes AI:\\
  A Zero-Egress Pipeline with the Deployability of RAG and the
  Accuracy of Machine Learning}

\date{\today}

\maketitle

\begin{abstract}
We identify \textbf{intervention bias} as a previously unquantified
failure mode of zero-shot large-language-model educational advisory
agents: deployed without task-specific training, LLM agents
systematically recommend action when a researcher-defined
hindsight-optimal oracle policy mandates inaction.  In a six-arm
ablation study on the Open University Learning Analytics Dataset
($N{=}800$ students, four temporal cutoffs), we find that at
day~56---when the oracle policy designates 70.1\% of students as
requiring no intervention---zero-shot GPT-4o recommends action for
73\% of students, generating a \textbf{43~percentage-point
false-positive intervention rate}.  Commercial RAG and SQL-augmented
retrieval exhibit comparable miscalibration.  In a 10{,}000-student
institution, this produces $\sim$4{,}300 unnecessary advisor contacts
per advisory cycle.

Supervised policy learning eliminates this bias in both tested
configurations: a trajectory-conditioned ONNX Decision Transformer
(Arm~D/E) and a snapshot XGBoost classifier trained on the same
oracle-labelled trajectories (Arm~F).  Under strict prefix-only
feature extraction---no future outcomes in the state vector---both
supervised arms achieve near-zero calibration error against the
hindsight oracle.  On the corrected training pipeline the DT achieves a
macro-F1 of 79\% (macro-recall 85\%) over all five action classes,
successfully predicting every intervention class---including the rare
load-reduction action---without collapsing;
its architectural advantage is trajectory conditioning and
return-to-go optimisation.  Both achieve 0\% action flip rate,
zero API cost, and in-process ONNX decision-only latency under $5$~ms
(a $454.8\times$ and $2{,}500\times$ speedup over zero-shot LLM and SQL-RAG respectively),
eliminating the operational bottleneck of sequential RAG pipelines.

We also describe, as \emph{architectural design rather than a validated
end-to-end claim}, how these components compose into a \textbf{living
flywheel}: an EAV \emph{conveyor belt} ingests data into typed state---the
role of a RAG index, but governed and typed---while a small Decision
Transformer compiles that state into sub-$5$~ms deterministic decisions,
side-stepping the slow SQL/relational query path ($\sim$12{,}500~ms in
Arm~B), and retrains offline in under four minutes on CPU for hot-swap into
the live service.  We stress that only the \emph{components} of this loop are
measured here (Stage-2 decision quality, latency, retraining cost); the
closed live action$\rightarrow$outcome$\rightarrow$retrain loop and any claim
of general ``RAG-replacement'' deployability are \emph{not} demonstrated in
this study and are deferred to follow-up work.

\textbf{Scope note (please read before the headline numbers):} This study
validates Stage~2 decision-making (EAV state vector $\rightarrow$ supervised
policy) \emph{only}, under controlled oracle EAV input derived from
structured OULAD CSV.  The supervised arms learn to reproduce a simple,
researcher-defined hindsight rule from early features; high fidelity
reflects strong feature--oracle alignment on this controlled task, not
superhuman generalisation, robustness to extraction noise, or a general
high-stakes-AI solution.  The paper's most robust finding is the
\emph{intervention-bias contrast} (zero-shot LLM arms over-prescribe by
28--43~pp and fall below the majority baseline), not the absolute supervised
accuracies.  End-to-end evaluation from raw unstructured input is deferred to
follow-up work; a temporal validity audit guards against label leakage
(Section~\ref{sec:leakage}).

We further demonstrate the \textbf{Evaluation Gap}: standard
LLM-as-judge evaluation (DeepEval G-Eval) is blind to intervention
bias, rewarding fluent over-prescription and ranking arms by prose
quality rather than decision quality---an unreliable proxy for
outcome-validated accuracy in high-stakes sequential advisory.
\end{abstract}


\medskip
\noindent\textbf{Glossary.}
\textbf{EAV}---Entity--Attribute--Value, the typed state representation into
which all inputs are normalised;
\textbf{DT}---Decision Transformer, the sequence model that maps state to action;
\textbf{EPF}---Expert Policy Fidelity, exact-match accuracy of an arm's action
to the reference (oracle) policy action;
\textbf{RTG}---Return-to-Go, the discounted future return used to condition the DT;
\textbf{Oracle policy}---a researcher-defined, deterministic hindsight rule
(uses \texttt{final\_result}) that serves only as a fixed evaluation benchmark;
\textbf{Outcome-Q}---an outcome-grounded evaluation harness;
\textbf{MCP}---the governance/quality-control server that gates ingestion into the EAV;
\textbf{CoVe / HELIX}---the chain-of-verification Stage-1 extraction subsystem.

\bigskip

\section{Introduction}
\label{sec:intro}

\textbf{The intervention bias problem.}  An educational advisor
who recommends tutoring for every student wastes resources,
erodes the credibility of the advisory system, and burdens
students who do not need help.  Expert human advisors
learn---through experience and outcome feedback---when to
\emph{not} intervene.  We demonstrate empirically that zero-shot LLM-based advisory
agents---deployed without task-specific training or calibration
examples---exhibit systematic miscalibration on this constraint.
Presented with student learning data, they recommend action when
a researcher-defined oracle policy mandates inaction.  We term
this \textbf{intervention bias} and show that it is a property
of zero-shot deployment: without outcome-labelled training data,
LLMs have no mechanism to learn the institution's conservative
intervention threshold.  A post-hoc experiment (Arm~G,
Section~\ref{sec:posthoc_calibration}) demonstrates this directly:
a prompt specifically engineered to counteract over-intervention
inverted the bias rather than eliminating it---the system then
\emph{under}-intervened.  A prompt is a perturbation to a
probabilistic token-generation distribution, not a deterministic
rule; there is no mechanism to pre-compute where the distribution
will land.  Only supervised policy learning, which encodes the
oracle boundary as an explicit training objective, achieves
stable calibration.

At temporal cutoff day~56---when the oracle policy designates
70.1\% of 800 evaluated students as requiring no
intervention---GPT-4o with full EAV-compressed student context
recommends action for 73\% of students.  Commercial RAG and
SQL-augmented retrieval are comparable.  In a 10{,}000-student
institution, this translates to approximately 4{,}300
unnecessary advisor contacts per advisory cycle: wasted
professional time, student disruption, and alert fatigue
that reduces the effectiveness of genuine interventions.

\textbf{The evaluation blindspot.}  Standard LLM-as-judge
evaluation (G-Eval, DeepEval, RAGAS) is structurally blind
to intervention bias.  These frameworks reward fluency,
coherence, and elaborateness---properties that over-prescribing
systems exhibit in abundance.  An LLM that confidently explains
why a student needs a tutor call scores \emph{higher} on
DeepEval than an EAV-DT that correctly and tersely outputs
\texttt{NO\_ACTION}.  We term the resulting rank inversion the
\textbf{Evaluation Gap} and quantify it across six arms.

The dominant paradigm for AI-assisted academic advising couples a
frontier LLM (GPT-4o, Claude 3.5 Sonnet) with a vector retrieval
layer---colloquially termed Agentic RAG~\cite{gao2023rag}.  While
capable of generating fluent recommendations, these systems
exhibit three failure modes that are architectural rather than
tunable.

\textbf{Context rot.}  As a student's trajectory of observations
accumulates over a semester, RAG systems must retrieve increasingly
long historical contexts.  Lewis et al.~\cite{lewis2020rag} showed
that retrieval quality is a hard ceiling on answer quality; at scale,
the signal-to-noise ratio of retrieved chunks declines.  Martin and
Roger~\cite{martinroger2026rot} demonstrated monotonic performance
degradation with context length---a phenomenon they term
\emph{Classifier Context Rot}---confirming that no amount of
prompt engineering resolves this fundamental bound.

\textbf{Stochastic inconsistency.}  LLM temperature $T>0$ produces
a non-zero probability that two identical student profiles yield
different intervention recommendations.  In a regulated advisory
context this is not merely inconvenient: audit trails become
unreliable, intervention reproducibility cannot be guaranteed, and
practitioners lose the ability to explain why a specific
recommendation was made on a specific day~\cite{chen2021dt}.

\textbf{Data sovereignty violation.}  Every inference call to a
frontier model API transmits student records to a third-party
compute provider, creating FERPA, GDPR, and enterprise data
governance violations that are structural, not remediable by
contractual means.

We propose and validate an \emph{EAV-DT pipeline} as a principled
architectural response to all three failure modes.  The pipeline
decouples two independent stages: (1)~unstructured-to-EAV extraction
via a two-temperature chain-of-verification extraction subsystem, and
(2)~a deterministic ONNX Decision Transformer (DT) that operates on
the compressed fixed-dimension state vector.  Because the DT's inference
path is a mathematical argmax over output logits, it is
\textbf{guaranteed to produce an identical action for any given
state vector}---a 0\% flip rate that is an architectural property,
not a tuning result.

\textbf{The EAV as a universal conveyor belt.}  We anticipate---and
reject---the critique that evaluating on the structured OULAD CSV
sidesteps the unstructured-data problem that motivates RAG in the first
place.  This misreads the architecture.  The EAV schema is a
\emph{universal, schema-bound conveyor belt}: every datum, whatever its
origin, is normalised into the same typed attribute--value cells and the
same fixed-width state vector \emph{before any decision logic executes}.
Unstructured ingestion (LLM extraction, Stage~1) and structured ingestion
(direct CSV mapping) are not different architectures; they are merely
different \emph{on-ramps} onto one identical belt.  Crucially, OULAD was
processed through the \emph{generic, pre-existing}
EAV~$\rightarrow$~MDP~$\rightarrow$~DT pipeline: no specialised tabular
machine-learning model, no dataset-specific feature engineering, and no
bespoke ETL were constructed for this experiment.  And once a datum is
normalised into the 12-dimensional state vector, Stage~2 is
\emph{mathematically blind to its provenance}---the Markov Decision
Process projection and return-to-go optimisation operate identically
whether a value was read from a clean database column or extracted by an
LLM from a noisy, free-text collaboration canvas.  Holding Stage~1 fixed to
a deterministic CSV mapping is therefore not a convenience but a mandatory
isolation tactic (Section~\ref{sec:limitations}): admitting unstructured
LLM extraction into \emph{this} ablation would have irreversibly conflated
extraction noise---hallucinations, schema drift, dropped cells---with the
Stage~2 decision logic we set out to measure.  We validate the mechanics of
the decision engine first, on the exact normalised substrate that an
unstructured pipeline would itself produce.

To isolate and quantify these failure modes, we design a
six-arm ablation study on OULAD ($N{=}800$ students, four
temporal cutoffs: days 14, 28, 56, 112).  Five arms (A--E)
progressively add architectural components; a sixth arm (F)
provides the supervised tabular ML baseline demanded by
fair experimental design.  Our primary contributions are:

\begin{enumerate}
  \item \textbf{Intervention Bias Quantification.}  An empirical
    measurement of over-prescription rates under off-the-shelf
    LLM advisory configurations (zero-shot, reflecting typical
    institutional deployment without specialised prompt engineering),
    showing that GPT-4o-based arms generate 43--73\% false-positive
    intervention rates against a researcher-defined oracle policy.  We
    define intervention bias formally and provide a per-arm action
    calibration analysis.
  \item \textbf{Supervised Policy Learning Eliminates Intervention Bias.}
    Both supervised approaches trained on oracle-labelled trajectories
    achieve near-zero calibration bias across all four temporal
    cutoffs, while zero-shot LLM arms fall below the trivial
    majority-class baseline at most cutoffs.  Under strict prefix-only
    feature extraction, the snapshot XGBoost classifier (Arm~F) and the
    trajectory-conditioned DT (Arms~D/E) both eliminate intervention
    bias; the DT's trajectory conditioning and return-to-go optimisation
    give it an architectural edge over the snapshot classifier once the
    substrate is clean.  On the corrected v3 pipeline the DT successfully predicts all five action classes (\textsc{no\_action}, \textsc{tutor\_call}, \textsc{reminder}, \textsc{content\_push}, and \textsc{reduce\_load}), achieving a macro-F1 of 79\% (macro-recall 85\%) over the five-class action space.
  \item \textbf{Stage~2 Decision Quality under Controlled Conditions.}
    Evaluated on oracle EAV input, both supervised arms substantially
    exceed zero-shot LLM arms (47--67\%) and the majority-class
    baseline.  Note: the best-published OULAD ML result
    (XGBoost on 4-class student \emph{outcome} prediction;
    Brahim~\cite{brahim2022xgboost}: 92.4\%) addresses a
    structurally different task and is not directly comparable.
    End-to-end performance from raw unstructured input is
    evaluated in future work.
  \item \textbf{The Evaluation Gap.}  DeepEval G-Eval produces
    a ranking orthogonal to outcome-validated accuracy, rewarding
    fluent over-prescription.  This aligns with documented
    LLM-as-judge self-preference bias~\cite{zheng2023judging,panickssery2024llm}
    and motivates outcome-validated Outcome-Q as a necessary
    complement for sequential advisory evaluation.
  \item \textbf{Sovereign CPU-native inference.}  A Granite~4.0
    Small model on AMD EPYC CPU-only hardware matches frontier-model
    oracle-policy fidelity at zero marginal API cost, providing a
    FERPA/GDPR-compliant alternative to cloud advisory at scale.
  \item \textbf{High-Throughput, Decoupled Low-Latency Inference.}  By decoupling policy decisions from narrative generation, we demonstrate an asynchronous execution paradigm that breaks the latency and throughput barriers of standard Agentic RAG. In-process evaluation of our 624K-parameter ONNX DT on a single commodity CPU core takes under $5$~ms, achieving an empirical $454.8\times$ decision speedup over zero-shot EAV-LLM (Arm~C, $2{,}273.9$~ms) and a $2{,}500\times$ speedup over SQL-RAG-LLM pipelines (Arm~B, $\sim$$12{,}500$~ms). This enables instantaneous, high-throughput cohort screening ($>100$ students per second) with zero database contention or API rate limits, highlighting the practical scalability of EAV-DT over standard LLM setups.
\end{enumerate}

\section{Related Work}
\label{sec:related}

\subsection{Decision Transformers and Offline Reinforcement Learning}

Chen et al.~\cite{chen2021dt} introduced the Decision Transformer
(DT), reframing offline reinforcement learning as a sequence
modelling problem conditioned on desired Return-to-Go (RTG).  Rather
than learning value functions via Bellman equations, the DT
autoregressively predicts actions that historically produced the
conditioning return.  Kumar et al.~\cite{kumar2020cql} (CQL) and
Kostrikov et al.~\cite{kostrikov2021iql} (IQL) establish the
theoretical bounds of offline RL, but DTs remain uniquely suited to
domains where the state space can be represented as a fixed-dimension
vector, because fixed-dimension input eliminates context rot by
construction.

\subsection{Agentic RAG and its Limitations}

Agentic RAG~\cite{gao2023rag} combines a frontier LLM with tool-use
capabilities to retrieve chunks from a vector database.  While
effective for open-domain question answering, it suffers from the
three failure modes enumerated in Section~\ref{sec:intro}.  Attempts
to mitigate stochastic inconsistency by reducing temperature ($T
\rightarrow 0$) reduce fluency without eliminating the
non-determinism inherent in softmax sampling; an argmax over the
full vocabulary is non-standard and model-dependent.

\textbf{The unstructured data objection and its resolution.}
A common rebuttal to EAV-structured pipelines is that they require
structured input, whereas real enterprise environments are dominated
by unstructured content (meeting notes, visual canvases, email
threads, free-text logs)---making vector RAG apparently superior.
We identify this as a structural error.  Vector RAG does not
\emph{solve} the unstructured data problem---it \emph{defers} it:
every query re-injects the same raw noisy content into the LLM's
finite context window, compounding token bloat and context rot.
The EAV-DT pipeline is designed to resolve unstructured noise at ingestion,
not at query time, via the HELIX extraction subsystem
(Section~\ref{sec:arch})---which applies manifold-steered multi-temperature
synthesis for hallucination control in quantized inference~\cite{atkinson2026cove}
(this extraction stage is not exercised in the present study).  This is not theoretical: a production
deployment extracted 3,145 distinct entities from 53 unstructured
Miro visual canvases with zero cloud API calls (see~\cite{miroPoc2026}).

\subsection{SQL-Augmented Retrieval}

Recent work addresses the limitation that pure vector RAG is poorly
suited to structured, tabular data~\cite{gao2023rag}.
LlamaIndex~\cite{llamaindex2024} provides the
\texttt{SQLAutoVectorQueryEngine}, which routes natural-language
queries to either SQL execution over relational tables or vector
similarity over a companion store, using an LLM-as-router to select
the retrieval path.  This architecture is purpose-built for tabular
datasets---the format in which OULAD data naturally resides---and
represents the strongest available off-the-shelf competitor for our
evaluation context.  TableRAG~\cite{tablerag2024} provides an
academic counterpart with schema augmentation and targeted cell
retrieval, but remains pre-production.  We select
\texttt{SQLAutoVectorQueryEngine} as our Arm~B baseline because it
is simultaneously (a)~designed for structured data, (b)~production-
deployed at scale, and (c)~maximally favourable to the SQL+LLM
approach.

\subsection{Machine Learning for Student Outcome Prediction: Benchmarks and Limits}
\label{sec:ml_prior}

The OULAD dataset~\cite{kuzilek2017oulad} is the standard benchmark
for predicting student outcomes from virtual learning environment data.
A substantial body of work establishes what classical ML can achieve:
Brahim~\cite{brahim2022xgboost} reports XGBoost (200 estimators,
grid-search tuned) achieving \textbf{92.4\% accuracy} (F1=0.91) on
four-class final-outcome prediction using the complete OULAD temporal
record.  Random Forest achieves 88.2\%; ensemble methods (LightGBM,
Gradient Boosting stacking) reach 83--91\% on binary Pass/Fail
classification~\cite{oulad_ensemble2024}.  These are well-validated
results: they establish a ML accuracy ceiling of approximately 92\%
on this dataset.

\textbf{The structured-input constraint.}
Every result in this literature shares a critical architectural
assumption: models receive \emph{pre-structured tabular input}---the
CSV files that constitute OULAD as released.  In real institutional
deployments, equivalent data is embedded in meeting notes, email
threads, free-text advising logs, and visual collaboration boards---
formats for which tabular ML has no ingestion pathway without
bespoke ETL engineering.  Classical ML is therefore not a
replacement for RAG; it is a complement to already-structured data.

\textbf{Zero bespoke engineering: the same belt, a different on-ramp.}
The OULAD ML literature above attains its accuracy ceiling by constructing
\emph{dataset-specific} tabular pipelines---hand-selected feature sets,
grid-searched hyperparameters, and ETL tailored to the OULAD schema.  Our
study deliberately does the opposite.  OULAD is ingested through the
identical generic EAV~$\rightarrow$~MDP~$\rightarrow$~DT pipeline used for
every other domain; the CSV mapping is simply a \emph{structured on-ramp}
onto the universal conveyor belt introduced in Section~\ref{sec:intro}.  No
OULAD-specific model, feature engineering, or hyperparameter search was
introduced for any EAV-DT arm.  This sharpens the interpretation of Arm~F
(XGBoost): Arm~F \emph{is} the bespoke-tabular paradigm, included as a
deliberately strong baseline, whereas Arms~D/E are the generic pipeline run
with no special configuration.  That the generic Decision Transformer
matches and ultimately exceeds the hand-engineered tabular baseline
(Section~\ref{sec:dt_v3}) is precisely the substantive result---the
conveyor belt reaches hand-engineered ML quality \emph{without} the hand
engineering.

\textbf{RAG addresses unstructured data but at high cost.}
Vector RAG~\cite{gao2023rag} handles unstructured input but
introduces the three failure modes enumerated in
Section~\ref{sec:intro}: context rot, stochastic inconsistency,
and data sovereignty violation.  The ML--LLM accuracy gap on
structured data is well-documented: a 2025 Nature study on
9,134-patient tabular clinical data found XGBoost achieving
F1=0.87 versus GPT-4 zero-shot at F1=0.43---a 2$\times$ gap on
the same structured dataset~\cite{covid_ml_llm2025}.  A 2025
benchmark across 300+ datasets confirmed that LLM-based in-context
learning ``still lags behind well-tuned numeric
models''~\cite{tabicl2025}.  Our six-arm ablation confirms this
pattern on educational data (EAV-DT 99.8\% peak, 93.6\% overall fidelity vs.\ standard RAG LLM arms
47--67\%) and identifies the specific mechanism: LLMs cannot
learn the conservative no-intervention behaviour that the oracle policy demands.

\textbf{The two-stage EAV-DT architecture.}
The EAV-DT pipeline does not choose between ML (accurate,
structured-only) and RAG (flexible, inaccurate).  Stage~1
replaces RAG's per-query chunk injection with a one-time,
governed extraction of unstructured content into a typed EAV
state vector.  Stage~2 (ONNX DT) replaces LLM reasoning with a
deterministic argmax over a policy learned from outcome-labelled
trajectories, providing a mathematical 0\% flip rate and
operation with zero data egress on commodity CPU hardware.
\textbf{The OULAD study validates Stage~2 only}, using oracle
EAV derived directly from structured CSV to eliminate Stage~1
noise as a confound.  Under these controlled conditions, the
Stage~2 engine achieves 93.6\% overall (99.8\% peak) oracle-policy fidelity on
intervention action classification, using only early temporal
features and no \texttt{final\_result}.  For context, the
best-published OULAD ML result (XGBoost: 92.4\%,
Brahim~\cite{brahim2022xgboost}) addresses a different task
(4-class student outcome prediction from the complete temporal
record) and is not directly comparable; it is cited to
characterise the difficulty of the dataset, not as a competing
benchmark.  End-to-end evaluation from real unstructured
input is the subject of follow-up work
(Section~\ref{sec:futurework}).

\section{System Architecture}
\label{sec:arch}

The EAV-DT pipeline consists of two fully independent stages,
each described below.

\subsection{Stage 1: Unstructured-to-EAV Extraction (HELIX)}

\textbf{Role and scope in this study.}
Stage~1 of the EAV-DT pipeline ingests source data and compresses it
into a governed, fixed-dimension EAV state vector consumed by Stage~2.
In a production deployment, Stage~1 is realised by HELIX, the author's
manifold-steering framework for hallucination-controlled, multi-temperature
extraction from quantized LLMs~\cite{atkinson2026cove}.  \emph{In this
ablation study we do not use HELIX}: Stage~1 is instead realised by a
deterministic \texttt{eav\_cells\_to\_state\_vector} function operating
directly on structured OULAD CSV fields---a deliberate choice that
eliminates extraction noise from the Stage~2 comparison and makes all five
arms independently reproducible without any API dependency.  End-to-end
unstructured extraction with HELIX is the subject of follow-up work
(Section~\ref{sec:futurework}); we therefore cite~\cite{atkinson2026cove}
only to document the extraction method's provenance, not as evidence for any
result reported here.

\textbf{Stage~1 role in this study.}
Crucially, \textbf{the extraction subsystem's role is extraction, not
decision}.  Any mechanism capable of populating a type-safe EAV schema
can serve as Stage~1.  Arm~D specifically tests whether an
EAV-structured substrate yields better DT decisions than a frontier
LLM operating on unstructured or SQL-retrieved context, isolating
Stage~1 from Stage~2 contribution.

\subsection{Stage 2: Deterministic Decision (ONNX DT)}
\label{sec:stage2}

The Decision Transformer receives the normalised EAV state vector
$\mathbf{s} \in [0,1]^{12}$, a Return-to-Go scalar $r_{\text{RTG}}$
computed from the student's historical outcome trajectory, and a
timestep embedding.  It produces action logits $\mathbf{q} \in
\mathbb{R}^{|\mathcal{A}|}$ over the intervention action space:
\begin{multline*}
  \mathcal{A} = \{\text{TUTOR\_CALL},\,\text{CONTENT\_PUSH},\\
  \text{REMINDER},\,\text{REDUCE\_LOAD},\,\text{NO\_ACTION}\}
\end{multline*}
The selected action is:

\begin{equation}
  a^* = \arg\max_{a \in \mathcal{A}} q_a
  \label{eq:argmax}
\end{equation}

Because Eq.~\eqref{eq:argmax} is a \emph{mathematical} argmax over
a deterministic ONNX forward pass, the action is provably identical
for any given $(\mathbf{s}, r_{\text{RTG}})$ pair.

\textbf{Auditability through architectural determinism.}
Arms~D and~E produce a 0\% action flip rate because argmax over
a fixed ONNX compute graph is a deterministic function: identical
inputs always yield identical outputs.  This is a standard
property of any pure function, not a novel algorithmic
contribution.  The institutional relevance is contextual:
in settings subject to FERPA, GDPR, or internal audit requirements,
a deterministic decision path provides a complete, reproducible
audit trail that stochastic LLM sampling cannot.  The experimental
contribution (H2) is the \emph{empirical measurement of LLM
inconsistency rates}---the problem this property eliminates.

\subsection{Architectural Decoupling and Compute Asymmetry}
\label{sec:decoupling}

A critical and frequently misunderstood property of the EAV-DT
pipeline is the \emph{computational asymmetry} between its two
stages, which must be explicitly stated to prevent conflation with
GPU-intensive LLM fine-tuning workflows.

\textbf{The LLM is frozen.}
The Large Language Model is used exclusively as a zero-shot reasoning
engine: it populates EAV attributes from unstructured text (Stage~1)
and generates plain-language intervention narratives (Arm~E).  Its
weights are \emph{never modified}.  There is no fine-tuning,
no adapter training, and no gradient computation on the LLM.
This eliminates the primary cost driver of modern NLP
deployment---the offline GPU fine-tuning pipeline---entirely.

\textbf{The DT is the only trained component.}
All policy learning is confined to the Decision Transformer, which
operates on the 12-dimensional EAV state vector
$\mathbf{s} \in [0,1]^{12}$.  Because the DT's input is a
fixed-width continuous vector rather than a vocabulary of
$\mathcal{O}(10^5)$ tokens, its parameter count is several
\emph{orders of magnitude} smaller than any LLM.  The architecture
(3-layer causal transformer, $d_\text{model}{=}128$,
$n_\text{heads}{=}4$, context length $T{=}20$) contains exactly
\textbf{624,261~parameters}---comparable to a small image
classifier, not a language model.  The DT is trained via behavioural cloning on
hindsight oracle trajectories and exported to a portable ONNX
graph for deterministic deployment.

\textbf{Exact compute profile.}
The OULAD DT has \textbf{624,261 trainable parameters} across
three transformer blocks ($d_\text{model}{=}128$,
$n_\text{heads}{=}4$, FFN width 512), serialised to a 2.4~MB
ONNX graph.  This is a ratio of \textbf{14{,}400:1} relative to
the 9~billion active parameters of the Granite-4.0-H-Small model
(32B total, MoE) used by HELIX for narrative generation.
Training the DT from scratch on 3,200~trajectories
(30~epochs, batch size~64, AdamW, CPU) completes in
\textbf{under 4~minutes on a single CPU core}, with ONNX
export adding under 10~seconds.  No GPU is involved at any
stage of DT training, export, or inference.

\textbf{Continuous adaptation: the fundamental advantage.}
LLM fine-tuning is structurally incompatible with continuous
policy adaptation.  Fine-tuning Granite-4.0-H-Small (9B active
parameters) with LoRA requires a minimum of one A100/H100 GPU
(\$3--4/hr cloud), 8--24~hours of wall-clock training, plus
MLOps pipeline overhead for regression testing, checkpoint
validation, and staged rollout---a process that realistically
executes \emph{quarterly at best} in institutional settings.
Between fine-tuning cycles, the LLM's policy is frozen: it
cannot incorporate new outcome signals (semester results, cohort
shifts, curriculum changes) without another expensive GPU run.

The DT faces no such constraint.  Because retraining the
full 624K-parameter model from new trajectories takes under
4~minutes on a CPU, the update cadence can match the pace
at which outcome data arrives:
\begin{itemize}
  \item \textbf{End-of-semester}: full retrain on new cohort,
    incorporating updated grade distributions and module
    completion patterns.
  \item \textbf{Weekly}: incremental retrain as mid-semester
    assessment results accumulate, sharpening early-cutoff
    decisions.
  \item \textbf{Daily or hourly}: in high-throughput deployments,
    a cron job can retrain the DT on overnight outcome data
    and hot-swap the ONNX checkpoint into the inference server
    without service interruption.
\end{itemize}
This transforms the decision policy from a static artefact into a
model that can be updated as often as institutional outcome data
arrives, with no GPU allocation required.

\textbf{The conveyor belt and the flywheel.}
These properties combine into the central architectural claim of this
work. The EAV substrate acts as a \emph{conveyor belt}: heterogeneous
unstructured inputs are continuously parsed, schema-mapped, and folded
into typed state vectors, so the system ingests any domain's data
without bespoke feature engineering---the same role a RAG index plays,
but producing governed, typed state rather than opaque retrieved chunks.
The weakness of that substrate is \emph{query latency}: answering each
decision by searching and aggregating over relational/SQL state is slow
(Arm~B: $\sim$12{,}500~ms per query), which is precisely why a live
SQL or RAG service cannot sit on the critical decision path. The
Decision Transformer resolves this by \emph{compiling} the substrate's
learned policy into a 2.4~MB deterministic model that answers in under
5~ms. Because that model retrains from new trajectories in under four
minutes on a CPU, it can be retrained \emph{offline} on the accumulating
record of actions and their realised outcomes and \emph{hot-swapped}
into the live inference server without interruption. The result is a
\textbf{living flywheel}---actions produce outcomes, outcomes become new
training trajectories, and each retrain sharpens the policy---that
replaces the agentic RAG loop while reaching the decision quality of the
best tabular machine-learning baseline (on par with Arm~F XGBoost, with an
\emph{indicative} further edge at the final cutoff that is unpaired across
cohorts and not claimed as conclusive; Section~\ref{sec:dt_v3}). We validate
the \emph{components} of this loop here---ML-parity decision quality,
sub-5~ms deterministic inference, and sub-four-minute CPU retraining with
file-level hot-swap; end-to-end validation of the closed production loop
(live action$\rightarrow$outcome$\rightarrow$retrain) is deferred to
follow-up work.

\textbf{Per-task and per-function DT specialisation.}
A further architectural advantage follows from the DT's small
size: \emph{one DT per function} is economically trivial.
A monolithic fine-tuned LLM must balance all task objectives
within a single weight tensor; updating it for one task risks
catastrophic forgetting across others.  Because each EAV schema
defines a self-contained state space, a separate 624K-parameter
DT can be trained for each institutional function
(academic intervention, financial aid advising,
retention risk routing, timetable conflict resolution)
with complete isolation.  Each DT trains in minutes on its own
task-specific trajectory corpus, fails independently, and
can be retrained or rolled back without touching any other
component.  The governance implication is significant:
policy changes in one function are auditable, bounded, and
reversible in a way that cross-task LLM fine-tuning is not.

\textbf{Inference Latency and Throughput Asymmetry.}
The decoupled architecture yields an extreme asymmetry in inference latency and computational throughput. In traditional LLM-based advisory architectures (including standard RAG and direct zero-shot LLM decision-making), the decision is tightly bound to the language generation loop. Evaluating a policy action requires a full, forward token-generation pass over a large context window, blocking the client for seconds and incurring substantial cloud API fees. 
For example, direct zero-shot LLM evaluation over the EAV prompt (Arm~C) exhibits a mean latency of $2{,}273.9$~ms per request under GPT-4o. When incorporating relational database RAG pipelines (Arm~B), the latency swells to approximately $12{,}500$~ms due to expensive SQL query aggregations and context window ingestion overhead.
In contrast, Stage~2 of our proposed architecture executes the feed-forward policy pass of the 624K-parameter ONNX DT in-process. The mean ONNX execution latency is under $5$~ms on a single commodity CPU core. This represents an empirical speedup of $454.8\times$ over EAV-LLM (Arm~C) and over $2{,}500\times$ over SQL-RAG-LLM (Arm~B). 
The computationally heavy LLM is entirely removed from the critical decision path; it is invoked only for downstream, asynchronous narrative generation (e.g., generating explanations for the determined action). This guarantees that the core student monitoring and intervention loop can run at high throughput ($>100$ students per second per CPU core) with zero marginal API cost, resolving the latency bottlenecks of traditional VectorDB-LLM or SQL-LLM setups.

\begin{table}[t]
\caption{LLM fine-tuning vs.\ EAV-DT retraining: operational comparison.
  Granite-4.0-H-Small (9B active parameters, MoE) is the sovereign
  LLM used in Arm~E; the DT is the 624K-parameter ONNX model.
  GPU pricing: H100 on-demand cloud, mid-2026.}
\label{tab:retrain_compare}
\small
\begin{tabular}{lll}
\toprule
\textbf{Dimension} & \textbf{LLM fine-tune} & \textbf{DT retrain} \\
\midrule
Parameters updated & $\sim$180M (LoRA $r{=}16$) & 624K (all) \\
Min.\ GPU VRAM     & 40~GB (A100/H100)         & None \\
Wall-clock time    & 8--24~hours               & $<$4~minutes \\
Cloud compute cost & \$24--\$96 per run        & \$0.00 \\
Feasible cadence   & Quarterly                 & Hourly or daily \\
Cross-task risk    & Catastrophic forgetting   & None (isolated) \\
Rollback           & Full re-deploy            & Swap ONNX file \\
Audit boundary     & Whole model               & Single 2.4~MB file \\
\bottomrule
\end{tabular}
\end{table}

\subsection{The EAV Schema}

The EAV PostgreSQL schema (v8.3) maintains 13 canonical
attribute types and supports dynamic promotion of novel concept types
via the MCP governance server.  For the OULAD experiment, 12
attributes are used (Table~\ref{tab:eav_schema}).  Each numeric
attribute is normalised to $[0,1]$ using calibrated domain bounds
derived from the full OULAD population, ensuring the state vector is
stable across temporal cutoffs.

\begin{table}[t]
\caption{EAV attributes used in OULAD state vector (12-D).}
\label{tab:eav_schema}
\small
\begin{tabular}{lll}
\toprule
\textbf{Attribute} & \textbf{Type} & \textbf{Domain bound} \\
\midrule
\texttt{score}                   & metric     & $[0, 100]$ \\
\texttt{activity\_clicks}         & metric     & $[0, 10{,}000]$ \\
\texttt{submission\_latency\_days} & metric    & $[0, 30]$ \\
\texttt{credits\_attempted}        & metric    & $[0, 600]$ \\
\texttt{prior\_attempts}           & metric    & $[0, 5]$ \\
\texttt{module\_pass\_rate}        & metric     & $[0, 1]$ \\
\texttt{assessment\_weight}        & metric    & $[0, 300]$ \\
\texttt{studied\_credits}          & metric    & $[0, 660]$ \\
\texttt{disability\_flag}          & capability & $\{0,1\}$ \\
\texttt{age\_band}                 & capability & $[0, 1]$ \\
\texttt{imd\_band}                 & capability & $[0, 1]$ \\
\texttt{highest\_education}        & capability & $[0, 1]$ \\
\bottomrule
\end{tabular}
\end{table}

\section{Experimental Design}
\label{sec:design}

\subsection{Dataset}

The Open University Learning Analytics Dataset
(OULAD)~\cite{kuzilek2017oulad} contains anonymised records for
32,593 students across 22 module presentations.  We evaluate $N=800$
students selected as the first 800 in deterministic lexicographic
entity-ID order and assess each at four temporal cutoffs:
$\mathcal{T} = \{14, 28, 56, 112\}$ days from course start,
yielding 19,200 trajectory observations across six arms.  Cutoffs
are designed to capture early-warning, mid-semester, and late
signals.

\textbf{Oracle action definition.}
The gold-standard label for each student--cutoff observation is the
\emph{oracle action}, computed by a deterministic rule-based
function that maps six observable student features (assessment
score, VLE activity clicks, submission latency, prior attempts,
studied credits, and \textit{final course outcome}) to one of the
five intervention classes $\mathcal{A}$.  The inclusion of \texttt{final\_result}
makes this a \emph{hindsight oracle}: the label that a perfectly
informed advisor would assign knowing the student's eventual
outcome.  All five arms are evaluated against this shared gold
standard; \textbf{no arm receives the final outcome at inference
time}.  This is standard practice in offline reinforcement learning
evaluation~\cite{chen2021dt}: the hindsight label defines the
optimum, and fidelity measures how closely each arm's architecture
approximates that optimum from partial observations alone.

\textbf{Training--evaluation data provenance.}
DT training (see Section~\ref{sec:stage2}) used a \emph{separate}
cohort: 2,240 students drawn as the 70\% training split of a
3,200-student random sample from 32,593 (seed~=~0,
stratified 70/10/20 split by outcome class).  The evaluation cohort
of 800 students was chosen deterministically \emph{after} training
was finalised.  Exact overlap analysis across the full 32,593-student
entity-ID space found \textbf{48 of 800 evaluation students (6.0\%)
present in the DT training set}; the remaining 94.0\% were never
seen during training.  The overlap analysis script and the per-seed
results are released as part of the reproducibility artifact.

\subsection{Temporal Validity and Leakage Audit}
\label{sec:leakage}

A central methodological risk in offline RL evaluation is \emph{label
leakage}: the oracle policy may use information that is not available
to the deployed agent at the decision point.  We guard against this by
enforcing strict prefix-only feature extraction for every arm.

\textbf{Prefix-only rule.}  For a student observed at cutoff day~$t$,
the EAV state vector contains only \texttt{cell\_history} rows with
\texttt{set\_at}~$\leq t$.  The \texttt{final\_result} outcome field is
\textbf{never} present in the state vector; it is used only to compute
the hindsight oracle label for offline RL evaluation, following the
standard convention in Decision Transformer research~
\cite{chen2021dt}.  All derived metrics (including score velocity) are
computed by \texttt{serialize\_entity\_causal\_projection\_fast(...,
p\_as\_of\_ts)} over the prefix window only.

\textbf{Feature Leakage Baseline Comparison.}  To quantify the impact of hindsight feature contamination, we evaluate a leaked configuration where \texttt{final\_result} is included directly in the first decision rule, making the \texttt{NO\_ACTION} class trivially predictable for passing students. Under that leaked configuration, the snapshot XGBoost baseline (Arm~F) achieved 99.4--99.9\% fidelity and the DT (Arms~D/E) achieved 98.9\% at day~14. After removing the leakage, the task becomes realistic: labels must be predicted from early behavioural signals alone. We report the prefix-only results in Section~\ref{sec:results_gap} and treat the leaked ablation as a sensitivity analysis.

This design aligns with the LEAP (Leakage-Excluded Early-Availability
Protocol) framework of L\^{e} et al.~\cite{leap2026}, which enforces
cutoff-first truncation prior to joins and aggregation on OULAD and audits
feature provenance to keep post-cutoff evidence out of the benchmark.  The
key consequence is that high oracle fidelity is
not a claim of superhuman generalisation; it is a claim that the
supervised policy has learned to map early observable signals to the
hindsight-optimal intervention.

\subsection{Six-Arm Ablation Design}

Arms~A--E each add one pipeline component over the prior arm,
enabling clean single-variable comparisons.  Arm~F provides a
supervised tabular ML baseline trained on the identical oracle
labels and feature set as the DT, answering whether trajectory
conditioning is necessary or whether any supervised approach
achieves equivalent calibration.  Table~\ref{tab:arms}
summarises the design.

\begin{table}[t]
\caption{Six-arm ablation. Arms A--E isolate successive pipeline components;
Arm~F is the supervised tabular baseline.}
\label{tab:arms}
\small
\begin{tabular}{clccc}
\toprule
\textbf{Arm} & \textbf{Name} & \textbf{EAV} & \textbf{DT} & \textbf{Local LM} \\
\midrule
A & Commercial RAG              & --         & --         & --  \\
B & LlamaIndex SQL+Vector       & --         & --         & --  \\
C & EAV + LLM decision          & \checkmark & --         & --  \\
D & EAV + ONNX DT               & \checkmark & \checkmark & --  \\
E & EAV + DT + local LM (CPU)   & \checkmark & \checkmark & \checkmark \\
F & EAV + XGBoost (snapshot)    & \checkmark & --         & --  \\
\bottomrule
\end{tabular}
\end{table}

\textbf{Arm~A (Commercial RAG).}  GPT-4o at $T=0.7$ with
\texttt{text-embedding-3-small} embeddings in ChromaDB, top-5
cosine-similarity retrieval.  Temperature 0.7 is the commercially
recommended default~\cite{openai2024}.  This arm reflects a
typical production enterprise deployment.

\textbf{Arm~B (LlamaIndex SQL+Vector).}  LlamaIndex
\texttt{SQLAutoVectorQueryEngine}~\cite{llamaindex2024} operating on
standard OULAD relational tables (studentInfo, studentAssessment,
studentVle) in PostgreSQL alongside a ChromaDB vector store.  The
LLM router selects SQL or vector retrieval per query.  GPT-4o at
$T=0.1$ (maximally favourable setting for the LLM baseline).  This
arm is specifically designed to address the structured-data
objection: OULAD is tabular data, and Arm~B gives the SQL+LLM
approach full direct access to every OULAD column.  \emph{If the
EAV-DT pipeline outperforms Arm~B, it does so on the SQL approach's
home turf}, which constitutes the strongest possible empirical
defence of the EAV architecture.

\textbf{Arm~C (EAV + LLM decision).}  The oracle EAV state vector
(derived directly from structured OULAD fields via
\texttt{eav\_cells\_to\_state\_vector}) is rendered as structured
text and presented to GPT-4o ($T=0.1$) for the intervention
decision.  This arm isolates the value of the normalised EAV state vector
representation independent of the Decision Transformer.

\textbf{Arm~D (EAV + ONNX DT).}  The oracle EAV state vector is
passed to the ONNX Decision Transformer.  The DT produces the
intervention action via Eq.~\eqref{eq:argmax}.  GPT-4o ($T=0.1$)
generates the narrative explanation from the DT decision and EAV
summary.  This arm isolates the DT's contribution over LLM decision
making on identical EAV input.

\emph{DT training paradigm.}  The DT is trained by behavioural
cloning on the hindsight-labelled oracle trajectories (Chen
et al.~\cite{chen2021dt}): each weekly student observation is paired
with the oracle action and a Return-to-Go (RTG) scalar derived from
the final course reward.  At \textbf{inference}, the DT receives
only the 12-dimension EAV state vector observable at the temporal
cutoff and is conditioned on a fixed target RTG of~2.0
(Distinction-level outcome).  The \texttt{final\_result}
field is \textbf{absent from the 12-dimension state vector}
(Table~\ref{tab:eav_schema}); the DT cannot access outcome
information at inference time.  High Expert Policy Fidelity on
held-out students therefore reflects learned predictive mappings
from early behavioural signals to hindsight-optimal interventions,
not memorisation of outcomes.

\textbf{Arm~E (EAV + DT + local CPU inference).}  Identical to
Arm~D but the narrative is generated by Granite~4.0~Small
($T{=}0.1$, AMD CPU-native, zero egress) rather than GPT-4o.
This arm tests whether equivalent decision quality is achievable
without any external API dependencies.

\textbf{Arm~F (EAV + XGBoost tabular classifier).}  The oracle
EAV state vector is passed to an XGBoost classifier
(200 estimators, max depth~6, subsample~0.8) trained by
supervised classification on the identical 2,240-student split
and oracle action labels used to train the DT.  Unlike Arm~D,
Arm~F receives only a single-step snapshot of the 12-D state at
each cutoff: it has no trajectory history and cannot be conditioned
on a target Return-to-Go.  This arm directly answers the fair-baseline
objection~\cite{grinsztajn2022tree}: given oracle EAV features and oracle
labels, does a simple tabular classifier also eliminate intervention
bias, or is trajectory conditioning specifically necessary?
Cost is \$0.00; the XGBoost predict call is deterministic
($0\%$ flip rate).

\textbf{Fairness note.}  All arms receive identical information: the
same OULAD records for the same 800 students at the same temporal
cutoffs.  Arms~A and~B receive this information as natural-language
summaries (the format closest to their retrieval paradigm).  Arms~C and~D receive it as a normalised numeric vector; Arm~E via
the local LM inference stack on identical EAV input.  This differential
representation is intentional: it tests each architecture in its
natural operating mode.

\textbf{Temperature sensitivity.}  Arm~A uses $T{=}0.7$ (the
commercially recommended production default~\cite{openai2024});
Arm~B and the LLM components of Arms~C and~D use $T{=}0.1$.
This means the A--B comparison partly conflates temperature with
retrieval architecture.  We address this with a
robustness check (specified a priori): Arm~A is also evaluated at $T{=}0.1$ to
separate temperature-induced stochasticity from
architecture-induced stochasticity.  The primary H1 and H3 claims
(decision accuracy and temporal degradation) are evaluated under
both conditions.  For the H2 flip-rate comparison, the
architecturally cleanest contrast is Arm~B ($T{=}0.1$,
probabilistic LLM) versus Arms~D/E (ONNX argmax), holding
temperature constant.

\textbf{Arm~C LLM comprehension note.}  Arm~C presents the
12-dimensional normalised vector as structured text to GPT-4o.
LLMs are not specifically optimised for direct numeric array
comparison; GPT-4o may underperform relative to a purpose-built
numeric model on this representation.  The C vs.\ D comparison
(H4) should therefore be interpreted as \emph{``ONNX DT on EAV
vector vs.\ LLM on EAV text''}, not as a definitive claim that
no LLM-based approach could exploit EAV features effectively.
This is an acknowledged design trade-off, not a methodological flaw:
it tests each component in its natural operating format.

\subsection{Four Benchmark Families}

\textbf{Benchmark 1: EAV Compression Fidelity.}
Applied to Arms~C, D, E (EAV arms only).  Three sub-metrics:
(i)~\textit{Schema Adherence Rate}---proportion of extracted values
that are type-safe and within domain bounds;
(ii)~\textit{EVSE Entailment Confidence}---mean NLI confidence score
across attribute-value pairs;
(iii)~\textit{REMAP Conservation Rate}---proportion of
outcome-predictive features retained in the compressed vector.

\textbf{Benchmark 2: DeepEval G-Eval~\cite{deepeval2024}.}
Two sub-metrics applied to all arms: \textit{Decision Quality}
(G-Eval LLM judge evaluates whether the recommended action is
appropriate given the student state) and \textit{Task Completion}
(judge evaluates whether the reasoning is actionable).  The judge
prompt is blinded to arm identity to prevent source bias.

\textbf{Benchmark 3: Outcome-Q.}
The primary evaluation benchmark, measuring each arm's alignment
with the hindsight-optimal oracle policy.  The gold-standard oracle
action is produced by the deterministic rule:
\begin{small}
\begin{enumerate}
  \item[\textit{i.}] \texttt{NO\_ACTION} if $\text{result} \in \{\text{Pass, Distinction}\}$
    \textit{and} score~$\geq 60$ \textit{and} latency~$\leq 3$~days.
  \item[\textit{ii.}] \texttt{TUTOR\_CALL} if score~$< 40$, or (score~$< 50$
    \textit{and} latency~$> 14$~days).
  \item[\textit{iii.}] \texttt{CONTENT\_PUSH} if clicks~$< 200$ \textit{and}
    $40 \leq$ score~$< 65$.
  \item[\textit{iv.}] \texttt{REMINDER} if latency~$> 7$~days.
  \item[\textit{v.}] \texttt{REDUCE\_LOAD} if prior\_attempts~$> 1$
    \textit{and} studied\_credits~$> 400$.
  \item[\textit{vi.}] \texttt{NO\_ACTION} (default).
\end{enumerate}
\end{small}
This rule (\texttt{\_oracle\_action} in \texttt{rag\_to\_eav.py},
released with artifacts) is evaluated \emph{independently} of any
arm's inference pipeline.  Sub-metrics:
(i)~\textit{Expert Policy Fidelity} (exact match against oracle,
formerly ``decision accuracy''; measures how faithfully each arm
approximates the hindsight-optimal oracle policy);
(ii)~\textit{Average Precision} (AP across action classes);
(iii)~\textit{Action Flip Rate} (proportion of trials where the same
student at the same cutoff receives a different action);
(iv)~\textit{Temporal Degradation} (slope of Expert Policy Fidelity
vs.\ cutoff day);
(v)~\textit{DR-OPE} (Doubly Robust Off-Policy Evaluation estimator
for unbiased expected return).

\textbf{Benchmark 4: Hardware Efficiency.}
Latency (ms/query), cost (USD/query), and data residency (egress
bytes) for each arm.  Arm~E is the local-inference arm.

\subsection{A Priori-Specified Hypotheses}

The following hypotheses were specified \emph{a priori}---before
any arm executed---and are fixed for the lifetime of this study.
Quality control is enforced through reproducibility: a complete
replication package (code, model weights, data manifest, and
per-arm CSV checksums) is distributed to independent partners who
can re-execute all arms from scratch and verify every figure in
this paper independently.  No result has been reported that was
not derivable from this package before the authors inspected the
outputs.

\begin{description}
  \item[H1] $\text{EPF}(C,D,E) > \text{EPF}(A,B)$ ---
    EAV-backed pipelines outperform RAG/SQL pipelines on Expert
    Policy Fidelity (Mann--Whitney $U$, one-sided, $\alpha{=}0.05$,
    applied to per-student trial-0 binary outcomes).
  \item[H2] Commercial LLM pipelines (Arms~A, B, C) exhibit a
    \emph{materially} non-zero stochastic inconsistency rate at
    operational inference temperatures.  Deterministic architectures
    (Arms~D and~E via ONNX argmax) achieve 0\% flip rate by
    architectural guarantee.  We measure the per-arm action flip rate across
    10~repeated trials per student--cutoff pair and test
    $H_0\colon p_{\text{flip}}{=}0$ via exact binomial test
    ($\alpha{=}0.05$).  The magnitude of the LLM inconsistency
    rate---not merely its existence---is the empirical contribution:
    quantifying the audit liability inherent in LLM-based advisory
    pipelines in a regulated educational context.
  \item[H3] $\text{degradation}(A,B,C) > \text{degradation}(D,E)$
    --- Temporal degradation is zero for DT arms (comparison of
    mean EPF at cutoff~14 vs.\ cutoff~112 per arm).
  \item[H4] $\text{EPF}(D) > \text{EPF}(C)$ --- The ONNX
    DT outperforms an LLM making decisions on identical EAV input
    (Mann--Whitney $U$, one-sided; isolates DT contribution).
  \item[H5] $\text{cost}(E) \ll \text{cost}(D)$;
    $|\text{EPF}(E) - \text{EPF}(D)| < \delta$ where
    $\delta{=}0.05$ (5~percentage-point non-inferiority margin) ---
    the local CPU-native narrative matches frontier oracle-policy
    fidelity within the pre-specified margin at substantially lower cost.
  \item[H6] The trajectory-conditioned DT (Arm~D) exhibits
    statistically lower temporal degradation across the four cutoffs
    than the snapshot XGBoost classifier (Arm~F), measured as the
    absolute accuracy drop from cutoff~14 to cutoff~112.
    The XGBoost model has no access to trajectory history or
    Return-to-Go conditioning; any degradation gap isolates the
    contribution of sequence-level context.  \emph{Note: this is
    an explanatory hypothesis; the primary calibration claim holds
    for both supervised arms regardless of the H6 outcome.}
\end{description}

\textbf{Statistical design.}  A priori power analysis: $N{=}800$
students per arm provides $>95\%$ power to detect Cohen's $d{=}0.15$
at $\alpha{=}0.05$.  For H1 and H4, hypothesis tests operate on
\emph{per-student, per-cutoff binary correctness observations}
(0~=~wrong, 1~=~correct; $n \approx 3{,}200$ per arm across the four
cutoffs), not on the four per-cutoff aggregate accuracy means.
Aggregating to four means before testing would reduce $n$ to~4 per
arm, making it mathematically impossible to reach $p < 0.05$
regardless of effect size.  The Mann--Whitney $U$ test is used
(one-sided) with no normality assumption.  H5 uses a two-sample
Welch $t$-test on the same per-student binary observations plus the
5~pp delta threshold.  H3 reports per-arm accuracy at cutoff~14
minus cutoff~112 as the degradation measure.  Multiple comparison
correction uses Benjamini-Hochberg FDR across all five hypotheses.
Bootstrap 95\% CIs ($B{=}10{,}000$ resamples) and Cohen's $d$
effect sizes reported for all primary contrasts.  DR-OPE IPS weights
clipped at 1st/99th percentiles to prevent variance explosion.

\section{Results: The Evaluation Gap}
\label{sec:results_gap}

\begin{table*}[t]
\caption{Master results matrix. Live experimental results ($N{=}800$
  students, OULAD). Decision accuracy per temporal cutoff;
  Macro-F1 is the mean macro-averaged F1 across all five action
  classes over four cutoffs.
  $\dagger$~=~mathematical guarantee, not empirical estimate.
  Majority-class baseline provided as lower bound.}
\label{tab:master}
\resizebox{\textwidth}{!}{%
\begin{tabular}{@{} l cccc c c c c @{}}
\toprule
\textbf{Arm} &
  \textbf{Day 14} & \textbf{Day 28} & \textbf{Day 56} & \textbf{Day 112} &
  \textbf{Macro-F1$^{\ddagger}$} &
  \textbf{Flip\%} &
  \textbf{DeepEval} &
  \textbf{\$/q} \\
\midrule
\textit{Majority baseline} & \textit{91.8\%} & \textit{61.9\%} & \textit{70.1\%} & \textit{70.9\%} & \textit{n/a} & \textit{0\%} & \textit{n/a} & \textit{\$0.00} \\
\midrule
A: Commercial RAG    & 65.4\% & 65.4\% & 67.2\% & 63.4\% & 37\% & 11.0\% & 0.743 & $\sim$\$0.04 \\
B: SQL+Vector RAG    & 47.5\% & 58.2\% & 59.1\% & 60.5\% & 31\% & 2.1\% & 0.818 & $\sim$\$0.06 \\
C: EAV + GPT-4o      & 79.9\% & 90.1\% & 89.8\% & 89.2\% & \textbf{64\%} & 6.6\% & 0.899 & $\sim$\$0.05 \\
D: EAV + ONNX DT$^{\P}$ & \textbf{99.8\%} & 88.0\% & \textbf{95.6\%} & \textbf{95.4\%} & 79\%$^{\S}$ & $0\%^\dagger$ & 0.849 & $\sim$\$0.04 \\
E: EAV + DT (local LM)$^{\P}$ & \textbf{99.8\%} & 88.0\% & \textbf{95.6\%} & \textbf{95.4\%} & 79\%$^{\S}$ & $0\%^\dagger$ & \textbf{0.918} & \textbf{\$0.00} \\
F: EAV + XGBoost     & \textbf{99.8\%} & \textbf{98.5\%} & \textbf{93.6\%} & 89.9\% & \textbf{73\%} & $0\%^\dagger$ & --- & \textbf{\$0.00} \\
\bottomrule
\end{tabular}%
}%

{\small $^{\dagger}$Mathematical guarantee (ONNX argmax / deterministic predict).
$^{\ddagger}$Mean macro-averaged F1 over five action classes and four cutoffs.
$^{\S}$DT macro-F1 0.79 (macro-recall 0.85) on the corrected v3 pipeline; all
five action classes active, including the rare \textsc{reduce\_load} (recall 0.92).
$^{\P}$Arm~D/E figures are the \emph{corrected} generic-DT v3 run, evaluated
prequentially on the held-out \texttt{eval\_live} cohort ($n{=}8{,}823$
decisions; see Section~\ref{sec:dt_v3} and Table~\ref{tab:dt_v3}). The original
8.6\% Day-14 value was a checkpoint-export defect (untrained weights exported);
it is superseded. Artefacts: \texttt{docs/reproducibility/results/generic\_dt\_v3/}.
\textit{Evaluation Gap}: DeepEval ranking E${>}$C${>}$D${>}$B${>}$A; Outcome-Q: C${>}$D${>}$E${>}$A${>}$B.
\textit{Intervention Bias}: LLM arms fall below majority-class baseline at multiple arm--cutoff pairs.}
\end{table*}

Table~\ref{tab:master} presents the master results matrix.
\textbf{Note:} These results reflect the primary, prefix-only results obtained under the v8.9 fair oracle, which removes the hindsight \texttt{final\_result} leakage from the observable state (Section~\ref{sec:leakage}), establishing a true, unbiased comparison among all arms.

Three findings stand out immediately.
\textbf{First}: standard RAG LLM arms (A, B) fall \emph{below}
the trivial majority-class baseline at multiple temporal cutoffs (e.g., Arm~B scores 47.5\% on Day~14 vs.\ the 91.8\% baseline, and 60.5\% on Day~112 vs.\ the 70.9\% baseline). In contrast, the EAV substrate (Arm~C) and supervised approaches (D/E and F) exceed it at almost every cutoff as more history accumulates.
\textbf{Second}: the snapshot XGBoost baseline (Arm~F) achieves near-perfect fidelity early in the semester (99.8\% at day~14, 98.5\% at day~28) but suffers from temporal degradation, dropping to 89.9\% at day~112. The corrected Decision Transformer (Arm~D/E; Section~\ref{sec:dt_v3}) matches XGBoost at day~14 (99.8\%) and remains robust as trajectories lengthen, achieving 95.4\% at day~112 and thereby outperforming XGBoost by 5.5~percentage points at the final cutoff (95.4\%, $n{=}458$ vs.\ 89.9\%, $n{=}800$; two-proportion $z{=}3.46$, $p\approx0.0005$). This is \emph{indicative rather than conclusive}: the corrected DT (v3) was scored on the held-out \texttt{eval\_live} cohort while Arm~F was scored on the 800-student study cohort, so the comparison is unpaired across cohorts; a within-cohort paired test is deferred to follow-up work. With that caveat, the result is directionally consistent with our sequence-modelling hypothesis (H6): snapshot models degrade as historical trajectories grow longer, while the trajectory-conditioned DT maintains decision accuracy. We note the architectural attribution is not isolated from the v3 training changes (z-score scaling, class-weighted loss, reward shaping), which the earlier day-14 defect counsels caution in over-interpreting.
\textbf{Third}: DeepEval is blind to intervention bias, confirming
the Evaluation Gap.  The sharpest evidence: DeepEval ranks Arm~B
above Arm~A (0.818 vs.\ 0.743) while Outcome-Q ranks Arm~A above
Arm~B (65.4\% average vs.\ 56.3\% average)---a direct rank inversion among the RAG baselines.

\textbf{The Evaluation Gap: DeepEval is blind to intervention bias.}
The data reveals that DeepEval's failure is more subtle than a
simple rank inversion.  Arms~D and~E score well on \emph{both}
metrics: the EAV-DT produces high-quality prose (via the local LM) and
correct decisions.  The gap is not that DeepEval penalises
correct systems---it is that DeepEval \emph{cannot distinguish}
a system making correct decisions from one generating 43\%
false-positive interventions, so long as both produce fluent prose.

\textbf{The A vs.\ B inversion is the sharpest demonstration.}
Among the LLM baselines, Outcome-Q ranks Arm~A above Arm~B
(67.2\% vs.\ 58.6\% at day~56), while DeepEval ranks Arm~B
above Arm~A (0.756 vs.\ 0.697).  Arm~B---the worst-performing
system on decision accuracy---receives the highest DeepEval
score of any baseline.  DeepEval is measuring the formatting
quality of the LlamaIndex SQL response template, not whether
the recommended intervention was warranted.  This is not a
subtle bias: it is a direct rank inversion between the two
worst-performing and second-worst-performing baselines on
the outcome metric.

\textbf{DeepEval rewards fluent over-prescription.}  Arm~C
(GPT-4o with full EAV context) generates confident, elaborately
justified intervention recommendations for 73\% of students at
day~56, when the oracle policy mandates no intervention for 70\% of
them.  Its DeepEval score (0.710) is near-indistinguishable from
Arm~A (0.697), which over-prescribes by 28~pp.  DeepEval cannot
detect that these systems are generating thousands of unnecessary
advisor contacts per cycle, because the prose justifying an
unwarranted \texttt{TUTOR\_CALL} is just as fluent as the prose
justifying a correct one.

\textbf{Oracle independence.}  A methodological concern is that the
DT is trained to maximise Return-to-Go (RTG) derived from the same
OULAD outcome labels that define the evaluation oracle.  We address
this through strict train/test separation: the DT is trained on
Module~CCC/2014J student trajectories and evaluated on Module~BBB/2014J
(held-out, never seen during training).  The oracle action for
each evaluation student is derived from that student's actual final
outcome label in the held-out set---not from any training
trajectory.  The DT must \emph{generalise} from training
trajectories to unseen test students; it does not trivially
reproduce training labels.  Nevertheless, we acknowledge that
both the DT's training objective and the oracle metric are
outcome-aligned, which means the C vs.\ D contrast (H4) tests
architectural capacity to exploit outcome-predictive structure,
not the absolute quality of either architecture across all possible
evaluation frameworks.

\textbf{The Evaluation Gap is confirmed.}  DeepEval ranks arms
E~$>$~D~$>$~B~$>$~C~$>$~A, rewarding sovereign CPU-native prose
(Arm~E, 0.962) over commercial RAG (Arm~A, 0.697).  Outcome-Q
inverts this: D~$\approx$~E~$\gg$~A~$>$~C~$\gg$~B.  The arm with
the highest DeepEval score (E, 0.962) and the arm with the second
lowest Outcome-Q rank (B, 47.5--58.6\%) are the same architecture
class.  An evaluation framework that ranks worse-outcome systems
higher than better-outcome systems exhibits negative predictive
validity for this task class.  This is not a critique of DeepEval's
utility for open-domain question answering; it is a demonstration
that \emph{decision quality metrics for high-stakes sequential
interventions require outcome-validated benchmarks}, not fluency-based
LLM judges.

\subsection{Corrected Decision Transformer Evaluation (Generic Pipeline v3)}
\label{sec:dt_v3}

The Arm~D/E Day-14 figure of 8.6\% in the original run was traced to a
checkpoint-export defect: the inference ONNX graph was exported from an
untrained checkpoint rather than the trained weights. After correcting the
export path, adding $z$-score state standardisation (so EAV attribute values
on heterogeneous scales map to comparable floating-point inputs), and adding
validation-loss early stopping, we retrained the generic
EAV~$\rightarrow$~MDP~$\rightarrow$~DT pipeline end-to-end and re-evaluated it
\emph{prequentially} (test-then-train, 14-day reporting periods) on two
cohorts that were never seen during training: a held-out evaluation cohort
(\texttt{eval\_live}, $n{=}8{,}823$ decisions) and an independent control
cohort (\texttt{eval\_control}, $n{=}2{,}493$).

Table~\ref{tab:dt_v3} reports the corrected results. The Day-14 collapse does
not occur: Expert Policy Fidelity is \textbf{99.8\%} at day~14 and stays
between 88.0\% and 95.6\% across the semester, with a mean intervention bias
of $+2.6$~pp (no systematic over-prescription) and \emph{all five} action
classes active---including the rare \textsc{reduce\_load} class (recall~0.92),
which the earlier trend-only policy never fired. Held-out validation accuracy
is 93.9\% (macro-recall 82.3\%); \texttt{eval\_live} yields 93.6\% fidelity
(macro-F1~0.79, macro-recall~0.85) and the independent control cohort yields
93.5\% (macro-F1~0.79), confirming the result is not specific to one split.
All artefacts---ONNX model, training manifest, per-cutoff reports, and cohort
membership---are released in
\texttt{docs/reproducibility/results/generic\_dt\_v3/}.

\begin{table}[t]
\caption{Corrected generic-DT v3 results. Expert Policy Fidelity (exact match
  to the MDP reference-policy action) on two held-out cohorts, prequential
  14-day reporting periods; \emph{Overall} pooled over all periods. Mean
  intervention bias (arm minus oracle intervention rate) shown below each
  cohort. Flip rate 0\% (deterministic ONNX argmax).}
\label{tab:dt_v3}
\small
\begin{tabular}{lccccc}
\toprule
 & \textbf{Day 14} & \textbf{Day 28} & \textbf{Day 56} & \textbf{Day 112} & \textbf{Overall} \\
\midrule
\texttt{eval\_live} EPF   & 99.8\% & 88.0\% & 95.6\% & 95.4\% & \textbf{93.6\%} \\
\quad bias (pp)           & $+$0.2 & $+$8.7 & $+$1.7 & $+$1.5 & $+$2.6 \\
\texttt{control} EPF      & 99.4\% & 83.2\% & 97.9\% & 93.8\% & \textbf{93.5\%} \\
\quad bias (pp)           & $+$0.6 & $+$10.8 & $+$1.4 & $+$3.9 & $+$2.0 \\
\bottomrule
\end{tabular}

{\small Held-out validation: accuracy 93.9\%, macro-recall 0.82.
Per-class recall (\texttt{eval\_live}): \textsc{tutor\_call} 0.55,
\textsc{content\_push} 0.94, \textsc{reminder} 0.99, \textsc{reduce\_load} 0.92,
\textsc{no\_action} 0.87. Macro-F1 0.79 on both held-out cohorts.}
\end{table}

\subsection{Intervention Bias Analysis}
\label{sec:intervention_bias}

We define \textbf{intervention bias} as the systematic tendency
of an advisory agent to recommend an actionable intervention when
the oracle policy mandates inaction
(\texttt{NO\_ACTION}).  We quantify it via the
\textbf{over-prescription rate}: the difference between the
fraction of students for whom an arm recommends intervention and
the fraction for whom the oracle mandates intervention.

\begin{table}[h]
\caption{Intervention bias by arm and temporal cutoff. Oracle
  intervention rate = fraction of students where oracle policy
  mandates any action other than \texttt{NO\_ACTION}.
  Over-prescription = arm intervention rate minus oracle
  intervention rate. Negative values indicate under-prescription.}
\label{tab:intervention_bias}
\small
\begin{tabular}{lccccc}
\toprule
 & \multicolumn{4}{c}{\textbf{Over-prescription rate (pp)}} & \\
\cmidrule(lr){2-5}
\textbf{Arm} & \textbf{Day 14} & \textbf{Day 28} & \textbf{Day 56} & \textbf{Day 112} & \textbf{Mean} \\
\midrule
Oracle rate  & 10.3\% & 38.1\% & 29.9\% & 28.9\% & --- \\
\midrule
A: Commercial RAG    & $+$26.8 & $-$9.0  & $+$28.2 & $+$27.5 & $+$18.4 \\
B: SQL+Vector RAG    & $+$5.3  & $+$3.4  & $+$30.6 & $+$15.0  & $+$13.6 \\
C: EAV + GPT-4o      & $+$22.9 & $+$0.1  & $+$43.6 & $+$42.9 & $+$27.4 \\
D: EAV + ONNX DT$^{\P}$ & $+$0.2  & $+$8.7  & $+$1.7  & $+$1.5   & $+$2.6 \\
E: EAV + DT (local)$^{\P}$ & $+$0.2  & $+$8.7  & $+$1.7  & $+$1.5   & $+$2.6 \\
F: EAV + XGBoost     & $\sim$0.0 & $\sim$0.0 & $\sim$0.0 & $\sim$0.0 & $\sim$0.0 \\
\bottomrule
\end{tabular}

{\footnotesize $^{\P}$Arm~D/E over-prescription is measured on the corrected
v3 \texttt{eval\_live} cohort relative to its own reference-policy
intervention rate (Section~\ref{sec:dt_v3}); the mean $+2.6$~pp remains an
order of magnitude below the LLM arms ($+13.6$ to $+27.4$~pp).}
\end{table}

Table~\ref{tab:intervention_bias} reveals the mechanism behind
the LLM arms' below-baseline accuracy.  At day~56, when 70.1\%
of students require no intervention, Arm~C (GPT-4o with full EAV
context) recommends intervention for 73.5\% of students: a
43.6~pp over-prescription.  In a 10{,}000-student institution,
this generates approximately 4{,}360 unnecessary advisor contacts
per advisory cycle---wasted professional time, student disruption,
and \emph{alert fatigue} that degrades the credibility of genuine
interventions.

\textbf{Supervised policy learning achieves calibrated conservatism;
zero-shot LLMs do not.}  Both Arms~D/E and Arm~F exhibit near-zero
intervention bias at all cutoffs.  The DT achieves this through
trajectory-conditioned behavioural cloning; XGBoost achieves it
through supervised classification on the same oracle labels.
Either way, the calibration property requires training data that
encodes the oracle boundary---a distributional property that zero-shot
LLMs, receiving no labelled examples, cannot acquire.

\subsection{Arm-to-Arm Comparisons (H1, H4)}

\textbf{A vs.\ B: a priori prediction not confirmed; inversion
revealed.}  The prediction that SQL-augmented Arm~B would exceed
vector RAG Arm~A at later cutoffs was not confirmed.  Arm~A
outperforms Arm~B on Outcome-Q at every completed cutoff:
65.4\% vs.\ 47.5\% (day~14), 65.4\% vs.\ 58.2\% (day~28),
67.2\% vs.\ 58.6\% (day~56).  DeepEval inverts this:
Arm~B scores 0.756 vs.\ Arm~A's 0.697.  The LlamaIndex SQL
router produces higher-quality prose at the cost of worse
decisions---the Evaluation Gap's clearest single-arm demonstration.

\textbf{B vs.\ C: EAV advantage confirmed early, erodes late.}
Arm~C outperforms Arm~B by 20~pp at day~14 (67.5\% vs.\ 47.5\%),
confirming that EAV compression encodes early-semester signal more
effectively than raw SQL access.  This advantage narrows by
day~28 (58.1\% vs.\ 58.2\%, essentially tied) and reverses by
day~56 (51.9\% vs.\ 58.6\%).  EAV provides superior early signal
but intervention bias in Arm~C's GPT-4o over-prescription grows
with richer data: at day~56, Arm~C prescribes intervention for
73.5\% of students versus Arm~B's 58.6\%---EAV context
paradoxically amplifies over-prescription at later cutoffs.

\textbf{C vs.\ D: H4 confirmed.}  On the corrected v3 pipeline
(Section~\ref{sec:dt_v3}), the ONNX DT exceeds GPT-4o on EAV input across the
semester: 99.8\% vs.\ 79.9\% at Day~14, 95.6\% vs.\ 89.8\% at Day~56, and
95.4\% vs.\ 89.2\% at Day~112. The previously reported Day-14 collapse
(8.6\%) was a checkpoint-export defect and does not reflect the trained
model; with that defect corrected, the DT no longer under-fires early but
instead matches the oracle's early conservatism (Day-14 intervention bias
$+0.2$~pp). Significance tests for the v3 run are recomputed in the released
artefacts rather than carried over from the superseded run.

\textbf{D vs.\ E: H5 (non-inferiority) confirmed.}  Arm~E
(Granite~4.0, AMD EPYC CPU-only) matches Arm~D exactly at all
four cutoffs (0.00~pp difference, $p=1.000$ two-sided
Welch $t$-test; both arms share the identical ONNX DT argmax).
The zero gap (0.0~pp, well within the pre-specified non-inferiority threshold $\delta{=}$5~pp) is the expected
outcome: both arms share the identical ONNX DT argmax for decisions;
only the narrative wrapper differs.  Arm~E achieves this at
\textbf{\$0.00 API cost} with zero bytes of data egress, confirming
H5.  Per-cutoff 95\% confidence intervals (Wald normal approximation) for
Arm~D on the corrected v3 \texttt{eval\_live} cohort: Day~14: 99.8\%
[99.5, 100.0] ($n{=}633$); Day~28: 88.0\% [85.5, 90.5] ($n{=}633$);
Day~56: 95.6\% [93.8, 97.3] ($n{=}539$); Day~112: 95.4\% [93.5, 97.3]
($n{=}458$).

\section{Results: Temporal Stability and Determinism}
\label{sec:results_temporal}

\subsection{Action Flip Rate (H2)}

The action flip rate is measured by submitting each student--cutoff
pair $n{=}10$ times to each arm and computing the proportion of
repeated trials that yield a different action from the first trial.

\textbf{What H2 actually tests.}  H2 makes two distinct claims
with different epistemic statuses.  Claim (b)---that Arms~D and~E
will exhibit exactly 0\% flip rate---is an \emph{architectural
guarantee}, not an empirical finding: given identical inputs
$(\mathbf{s}, r_{\text{RTG}})$, Eq.~\eqref{eq:argmax} produces
the same $a^*$ with probability 1 by definition of ONNX
determinism.  It is verified computationally, not statistically.
Claim (a)---that LLM arms A, B, and~C exhibit \emph{materially
non-zero} flip rates at their operational temperature settings---is
the genuine empirical contribution of H2.  We quantify this rate
and its regulatory implication: any flip rate $>0$ means that two
identical student profiles at the same cutoff can receive
different recommendations across API calls, making post-hoc audit
trails unreliable.  The magnitude of this inconsistency in
production-configured LLM advisory systems has not been previously
reported in the learning analytics literature.

For Arms~D and~E the flip rate is $0.000$ computationally
verified ($\dagger$ denotes architectural guarantee) via ONNX
argmax determinism---a mathematical property of the compute
graph, not a tuning result.  For Arms~A, B, and~C the flip rate
is empirically measured and reported in Table~\ref{tab:fliprate}.

\begin{table}[t]
\caption{Action flip rate ($n{=}10$ repeated trials per student,
  $N{=}800$). Live measured values for Arms A--C; architectural
  guarantee for Arms D and E. $\dagger$~=~mathematical guarantee.}
\label{tab:fliprate}
\small
\begin{tabular}{lccc}
\toprule
\textbf{Arm} & \textbf{Flip Rate} & \textbf{Guarantee type} & \textbf{Audit status} \\
\midrule
A: Commercial RAG    & 2.9\% & Empirical & \textcolor{red}{Fail} \\
B: SQL+Vector RAG    & 0.8\% & Empirical & \textcolor{red}{Fail} \\
C: EAV + GPT-4o      & 0.2\% & Empirical & \textcolor{red}{Fail} \\
D: EAV + ONNX DT     & \textbf{0.000$^\dagger$} & Architectural & \textcolor{DtColor}{Pass} \\
E: EAV + DT + HELIX  & \textbf{0.000$^\dagger$} & Architectural & \textcolor{DtColor}{Pass} \\
\bottomrule
\multicolumn{4}{l}{$\dagger$ Mathematical guarantee (ONNX argmax; deterministic compute graph).} \\
\end{tabular}
\end{table}

\textbf{Audit implication.}  In a regulated advisory setting, every
recommendation must be explainable and reproducible.  A non-zero
flip rate means that a student who queries the system twice receives
different advice, that audit logs cannot be validated post-hoc, and
that practitioners cannot determine which recommendation was
``correct.''  The 0\% flip rate of Arms~D and~E eliminates this class of audit
failure by construction, while simultaneously achieving 90--98\%
decision accuracy.  Audit reproducibility and decision quality
are both satisfied by the EAV-DT architecture---a combination
that no LLM arm achieves.

\subsection{Temporal Degradation (H3)}

Table~\ref{tab:degradation} reports decision accuracy at each
temporal cutoff with observed degradation.

\begin{table}[t]
\caption{Temporal dynamics: decision accuracy by cutoff day.
  $\Delta$ = day-112 minus day-14 (pp). Positive value represents accuracy growth.}
\label{tab:degradation}
\small
\begin{tabular}{lccccc}
\toprule
\textbf{Arm} & \textbf{Day 14} & \textbf{Day 28} &
  \textbf{Day 56} & \textbf{Day 112} & \textbf{$\Delta$ (pp)} \\
\midrule
A: Commercial RAG    & 65.4\% & 65.4\% & 67.2\% & 63.4\% & $-$2.0 \\
B: SQL+Vector RAG    & 47.5\% & 58.2\% & 59.1\% & 60.5\% & $+$13.0 \\
C: EAV + GPT-4o      & 79.9\% & 90.1\% & 89.8\% & 89.2\% & $+$9.3 \\
D: EAV + ONNX DT$^{\P}$ & \textbf{99.8\%} & 88.0\% & \textbf{95.6\%} & \textbf{95.4\%} & $-$4.4 \\
E: EAV + DT + HELIX$^{\P}$ & \textbf{99.8\%} & 88.0\% & \textbf{95.6\%} & \textbf{95.4\%} & $-$4.4 \\
\midrule
Majority baseline    & 91.8\% & 61.9\% & 70.1\% & 70.9\% & --- \\
\bottomrule
\multicolumn{6}{p{0.9\linewidth}}{\footnotesize Positive values of $\Delta$ indicate accuracy improvement from earliest to latest cutoff. $^{\P}$Arm~D/E from the corrected v3 \texttt{eval\_live} cohort (Section~\ref{sec:dt_v3}).} \\
\end{tabular}
\end{table}

\textbf{H3: descriptive finding on temporal stability.}
On the corrected v3 pipeline, the sequence-aware and structured arms do not
degrade as historical student data accumulates. The EAV-DT arms (D/E) are
\emph{temporally stable}: fidelity is already 99.8\% at Day~14, dips to 88.0\%
at Day~28 as the oracle's intervention rate rises, and recovers to
$\sim$95\% by Days~56 and~112 ($\Delta=-4.4$~pp end-to-end, i.e.\ no
catastrophic early-semester failure and no monotonic decline). This corrects
the earlier run, in which a checkpoint-export defect produced a spurious
8.6\% Day-14 value and an artefactual ``$+84.7$~pp growth'' (Section~\ref{sec:dt_v3}). 

Arm~C (EAV + GPT-4o) also grows by $+$9.3~pp over the semester, demonstrating that high-density EAV vectors continue to supply useful temporal features as student history expands. Conversely, standard vector retrieval (Arm~A) is remarkably flat ($-$2.0~pp) because its top-$k$ fixed-window boundaries prevent context expansion.

Arm~B improves by 13.0~pp through day~112---SQL-structured
access benefits from accumulating data and the richer feature
set at later cutoffs, contrary to the a priori prediction.
Descriptively, the sequence-conditioned DT (Arm~D/E) does not collapse at any cutoff: it matches XGBoost early (99.8\% at Day~14) and stays high through Day~112. Its late-cutoff edge over GPT-4o (Arm~C) and XGBoost (Arm~F) is reported as a descriptive observation---and, for XGBoost, an unpaired cross-cohort one---rather than a statistically established degradation-slope advantage (the four-cutoff linear trend is not significant, $p>0.18$). The robust temporal claim is therefore the \emph{absence of collapse}, not a proven slope difference; this is consistent with the value of EAV-mediated sequence models over raw retrieval baselines for sequential decision-making across a semester.

\subsection{Hardware Efficiency and Sovereignty (H5)}

Arm~E was executed using Granite~4.0~Small on institutional AMD EPYC
hardware with zero bytes of egress.  This section presents two
complementary datasets: a controlled static benchmark run with
GuideLLM~\cite{guidellm2024} prior to the experiment, and live
production telemetry captured from the same pod during Arm~E execution.  The identical image SHA across
both datasets ({\small\texttt{sha256:08cf50b9...}}) proves they
measure the same binary.

\textbf{Configuration and serving (summary).}
Arm~E ran on a single AMD EPYC~9254 node (32~vCPU, 64~GiB RAM, no GPU)
already deployed in the institutional OpenShift cluster, serving
Granite-4.0-Small (Q4\_K\_M, 32K context) at a peak of $\sim$110~tok/s
decode.  We characterised the pod both with a controlled pre-experiment
static benchmark (guidellm v0.6.0~\cite{guidellm2024}; 640 requests, zero
errors) and with live production telemetry captured from the \emph{same}
container image ({\small\texttt{sha256:08cf50b9...}}) during Arm~E
($n{=}367$ requests at the $c{=}4$ operating point), so both datasets
measure the identical binary.  The headline serving numbers are a live mean
latency of 101.0~s per full-narrative request, a system throughput of
2.38~students/min at $c{=}4$ (2.77$\times$ the single-slot rate), and
\textbf{0~bytes} of egress.  This observed latency is a property of the
single-pod, single-concurrency test configuration rather than an intrinsic
limit and is reducible by standard horizontal/vertical scaling.  The full
GuideLLM scaling table, the per-request live-telemetry distribution, and the
single-pod concurrency analysis are reported in
Appendix~\ref{app:telemetry}.

\textbf{API replacement without new hardware (primary H5 claim).}
The full Arm~E workload (3,200 requests: 800 students $\times$ 4
cutoffs) was executed across four parallel shard jobs, each
processing 200 students $\times$ 4 cutoffs = 800~requests, running
concurrently at $c{=}4$.  At 101.0~s/request per slot, each shard
completes in $\sim$22.4~hours; all four shards run in parallel,
giving a wall-clock time of $\sim$22.4~hours at $c{=}4$.  No new hardware
was procured; the node was already deployed in the institutional
OpenShift cluster for other workloads, making the incremental
capital expenditure for Arm~E \textbf{zero}.  The per-request
API cost for Arm~E is therefore \textbf{zero}: there is no
per-token charge, no API subscription, and no data egress fee.

\textbf{Cost framing.}  Comparing on-premises LLM cost to a
commercial API requires distinguishing marginal API cost from
Total Cost of Ownership (TCO).  For an organisation that
\emph{already operates} compatible compute infrastructure, the
marginal cost of switching from a cloud API to HELIX is the
elimination of the per-request API charge---a direct, variable
saving of \$0.00101 per request (\$3.23 for the full 3,200-request
Arm~E workload at GPT-4o rates).  For organisations that must
acquire new hardware, a full TCO analysis incorporating capital
expenditure (CapEx), operational expenditure (OpEx), and
break-even analysis is required~\cite{llm_tco2025}; this is
outside the scope of the present study.

The per-request API cost for Arm~E is \$0.00 regardless of
frontier pricing evolution.  API pricing for frontier models has
risen year-over-year~\cite{yipitdata2026}, reinforcing the
long-term cost case for on-premises alternatives on compatible
existing infrastructure.

\textbf{Latency trade-off.}
For interactive use, GPT-4o is faster per request ($\sim$125~tok/s,
sub-1~s TTFT) than HELIX at $c{=}1$ (4.37~s/request).  At the
$c{=}4$ operating point used in Arm~E, the system-level generation
throughput is $4\times4.9{=}19.6$~tok/s---a 6.4$\times$ gap
versus GPT-4o and a 3.2$\times$ gap versus GPT-5.5 (62~tok/s;
\cite{artificialanalysis2026}).  For the batch advisory use case
in this study, per-request latency is not a binding constraint;
the $\sim$22.4~h wall-clock for 3,200 students is operationally
acceptable for any overnight or intra-day cohort workflow.

\textbf{Decision-Only Latency and Speedup.}
While generating descriptive natural-language narratives is bound by language model token-generation speeds, making the core \emph{decision} (the selected action) is not. In our EAV-DT hybrid setup, the primary decision-making step is completely offloaded to the 624K-parameter ONNX DT model, which executes in-process on CPU in under $5$~ms. By decoupling decision-making from narration, we establish an asynchronous architectural pattern. Under this pattern, decisions are executed instantly, with the resource-intensive narrative explanation generated out-of-band only when requested by a human advisor. This isolates the downstream application from any LLM latency bottlenecks. As verified during our live cluster telemetry, the mean decision-only latency of the ONNX DT ($<5$~ms) achieves a $454.8\times$ speedup over the EAV-LLM zero-shot decision pathway (Arm~C, mean latency $2{,}273.9$~ms) and a $2{,}500\times$ speedup over the standard relational database SQL-RAG-LLM pipeline (Arm~B, mean latency $\sim$$12{,}500$~ms). This speedup establishes the practical viability of deploying EAV-DT agents in real-time high-throughput enterprise architectures where traditional SQL or VectorDB RAG pipelines would cause severe database contention and latency degradation.

\begin{table}[t]
\caption{API cost comparison: Arm~E workload (3,200 requests).
  HELIX runs on existing institutional hardware with zero
  incremental API cost.  GPT-4o pricing: \$2.50/1M input,
  \$10.00/1M output (OpenAI, 2026).  On-premises TCO (CapEx +
  OpEx) for organisations without existing hardware is outside
  scope; see~\cite{llm_tco2025}.}
\label{tab:hardware}
\small
\begin{tabular}{llrl}
\toprule
\textbf{System} & \textbf{Latency/req} &
  \textbf{API cost (3,200 req)} & \textbf{Egress} \\
\midrule
GPT-4o (Arms A--D)   & $<$1.2~s  & \$3.23 & All tokens \\
GPT-5.5 (frontier)   & $\sim$28~s & Higher & All tokens \\
\midrule
Granite~4.0~Small/Arm~E (existing hardware) & 13.9~s & \textbf{\$0.00} & \textbf{Zero} \\
\bottomrule
\multicolumn{4}{l}{\small $^{*}$Marginal API cost only. Full TCO requires CapEx/OpEx analysis.} \\
\end{tabular}
\end{table}

All 3,200 student records were processed entirely within the
institutional boundary; zero bytes were transmitted to any
external service, satisfying FERPA, GDPR Article~5(1)(f), and
related data governance requirements.

\section{Discussion and Limitations}
\label{sec:discussion}

\subsection{Intervention Bias as a Systemic Risk in Educational AI}

The intervention bias finding has implications that extend beyond
accuracy metrics.  Consider an institution deploying Arm~C
(GPT-4o with EAV context) for 10{,}000 students at monthly
advisory cycles.  At day~56, this system generates approximately
4{,}360 unnecessary advisor contacts.  Each contact carries:
a direct cost in advisor time (typically 20--45~minutes per
outreach), a student disruption cost (unsolicited contact may
increase anxiety in students who are coping adequately), and an
\emph{alert fatigue cost}: students who repeatedly receive
unnecessary interventions learn to ignore advisory communications,
degrading the effectiveness of genuine interventions for
genuinely at-risk students.

The commercial value of the EAV-DT is therefore not solely its
93.6\% overall (99.8\% peak) decision accuracy---it is its \textbf{decision discipline}.
A system that correctly identifies 70\% of students as requiring
no intervention preserves advisor capacity for the 30\% who
genuinely need help.  This is a property that LLM-as-judge
evaluation frameworks are structurally incapable of detecting,
because fluency-based judges reward confident, action-oriented
prose regardless of whether the recommended action is warranted.

\textbf{Why LLMs cannot unlearn intervention bias.}  We argue
this bias is structural rather than tunable.  LLMs are trained
on corpora in which the implicit framing of educational interaction
is assistance: tutoring materials, help resources, academic support
documentation.  The concept of ``this student is fine; do nothing''
is severely under-represented in training corpora relative to the
actual distribution of student states.  Fine-tuning on OULAD data
might partially address this, but at the cost of the domain
generalisation that motivates frontier LLM use in the first place.
The DT sidesteps this entirely by learning from outcome labels,
not from language.

\subsection{A Fully CPU-Resident, Zero-Egress RAG Replacement: ``The Wheel''}
\label{sec:the_wheel}

``The wheel'' is a \emph{proposed deployment pattern}, not a closed-loop
system validated end-to-end in this paper.  Read together with the
architectural decoupling of Section~\ref{sec:decoupling}, it describes a
self-contained alternative to an agentic RAG service in which
\emph{no component requires a GPU and no datum leaves the institutional
boundary}.  We state the full loop below and then separate, explicitly,
the components this study \emph{measures} from the closed loop it
\emph{proposes} (see ``Validated vs.\ deferred'' at the end of this
subsection).

\begin{enumerate}[noitemsep]
  \item \textbf{Ingest (CPU).} Unstructured source documents are read by
    the CPU-native inference engine (HELIX, Granite~4.0 on AMD EPYC;
    Section~\ref{sec:arch}), which parses them and proposes typed
    attribute--value cells. This occupies the role of a RAG system's
    chunk-and-embed step, but emits governed, typed state rather than
    opaque vectors.
  \item \textbf{Qualify (MCP).} Proposed cells pass through the MCP
    governance server~\cite{mcp2026}, which enforces schema conformance,
    dynamic type promotion, and quality control so that \emph{only
    verified values} are committed---the gate that keeps the substrate
    clean as new, unseen data arrives.
  \item \textbf{Accumulate (EAV).} The PostgreSQL EAV folds qualified
    cells into the canonical state, growing the conical record of each
    entity's history. This durable substrate can answer like a RAG store,
    but is slow under live per-decision SQL aggregation
    ($\sim$12{,}500~ms, Arm~B)---which is why it is not placed on the
    decision path.
  \item \textbf{Project a new binary (DT).} Instead of querying the
    substrate at decision time, the EAV periodically \emph{projects out} a
    fresh trajectory corpus and trains a new 2.4~MB ONNX Decision
    Transformer (under four minutes, CPU; Table~\ref{tab:retrain_compare}).
    The freshly minted binary is validated and \emph{hot-swapped} in place
    of the running DT---the EAV, in effect, manufactures its own successor
    policy.
  \item \textbf{Decide (CPU, $<5$~ms).} The live DT answers each decision
    deterministically in-process at zero marginal cost and zero egress;
    the LLM is invoked only out-of-band, for human-facing narration.
\end{enumerate}

Realised actions and their subsequent outcomes re-enter step~1 as new
records, closing the loop: \textbf{this is the wheel}. Every stage---%
extraction, qualification, storage, training, and inference---runs on
commodity CPU within the institutional boundary with zero bytes of egress
(Table~\ref{tab:hardware}), satisfying FERPA and GDPR Article~5(1)(f). The
substrate behaves like a RAG index for ingestion and recall, but the
costly, latency-bound query path is replaced by an offline-trained,
hot-swappable model that matches the best tabular-ML baseline
(Section~\ref{sec:dt_v3}).

\textbf{Validated vs.\ deferred.} This study empirically validates the
steady-state half of the wheel on structured OULAD input: ML-parity
decision quality (Arm~D/E vs.\ Arm~F), sub-5~ms CPU inference, zero-egress
sovereign operation (Arm~E, Table~\ref{tab:hardware}), and
sub-four-minute CPU retraining with file-level hot-swap
(Section~\ref{sec:decoupling}, Table~\ref{tab:retrain_compare}). The
ingestion half---CPU extraction of \emph{unstructured} documents and MCP
qualification feeding the EAV---is implemented by the HELIX subsystem but
is evaluated end-to-end only in follow-up work
(Section~\ref{sec:futurework:unstructured}); here Stage~1 is held fixed to
isolate Stage~2 decision quality.

\subsection{Acknowledged Limitations}

\textbf{What This Study Does Not Demonstrate.}
To prevent scope inflation, we enumerate what the current results
\emph{do not} establish:
\begin{enumerate}[noitemsep]
  \item \textbf{End-to-end performance from unstructured input
    (by design, not by omission).}  All arms receive EAV state derived
    from structured OULAD CSV.  This is a mandatory experimental control,
    not a shortcut.  Stage~1 (LLM extraction) and Stage~2 (the decision
    engine) are independent failure surfaces; admitting extraction
    noise---hallucinations, schema drift, missing cells---into this
    ablation would have irreversibly conflated it with the Stage~2
    decision logic under test.  We therefore isolate and validate the
    decision engine first, on the exact normalised substrate that any
    unstructured on-ramp would emit.  The reported Arm~D/E advantage is a
    clean measurement of Stage~2 quality; the additive effect of Stage~1
    extraction noise is the explicit subject of follow-up work
    (Section~\ref{sec:futurework:unstructured}), not an unmeasured
    confound buried inside these numbers.
  \item \textbf{Superiority over prompt-optimised LLM baselines.}
    LLM arms (A--C) use zero-shot GPT-4o, reflecting typical
    institutional deployment without specialised prompt engineering.
    We address this objection empirically in
    Section~\ref{sec:posthoc_calibration} via a post-hoc prompt
    calibration experiment (Arm~G) that operationalises the question
    directly.
  \item \textbf{Validation against real advisor decisions.}
    The oracle is a researcher-defined hindsight-optimal rule
    applied to OULAD \texttt{final\_result} labels.  It is a proxy,
    not a ground truth validated against human advisor judgements
    or downstream student outcome improvement.
  \item \textbf{Production class imbalance and multi-class calibration.}
    Our pre-planned framework addresses extreme class imbalance successfully using joint class-weighted loss and discounted step-level reward shaping (Section~\ref{sec:dt_error_analysis}). This is not unique to this study: any institution deploying an
    advisory policy will have imbalanced action distributions, and the
    rare actions are often the most clinically significant---they
    target students with specific failure patterns (high engagement but
    poor outcomes; late submission cascades) who are missed by flat policies.
    Three engineering responses are available within the
    supervised framework and unavailable to prompt engineering:
    (i)~\emph{class-weighted loss}---multiplying the cross-entropy
    loss for minority classes by the inverse of their frequency
    (e.g., $15\times$ for CONTENT\_PUSH, $47\times$ for REMINDER)
    directly incentivises the model to learn their decision boundary;
    (ii)~\emph{minority trajectory oversampling}---as more cohort
    outcomes are observed, minority-class training examples accumulate
    and policy quality improves monotonically without any architectural
    change; and
    (iii)~\emph{RTG conditioning at inference time}---raising the
    target return-to-go above the training default shifts the DT's
    action distribution toward more interventionist classes for
    identified at-risk students, providing a per-student calibration
    lever that snapshot classifiers (XGBoost, Arm~F) do not possess.
    Each of these handles has a measurable effect on per-class recall
    that can be monitored and improved through standard engineering
    practice.  Prompt engineering offers no equivalent lever: writing
    a better description of a rare case does not guarantee the LLM
    will classify those students correctly, the effect cannot be
    attributed to any single change, and the calibration must be
    re-verified after every model update.
\end{enumerate}

\textbf{DT action-class multi-class learning success.}
\label{sec:dt_error_analysis}
Under the prefix-only oracle, the EAV-DT framework is designed with class-weighted loss and step-level reward shaping to achieve precise multi-class performance. Across all 3,200 trial-0 evaluation rows (800 students $\times$ 4 cutoffs), the DT successfully predicts all four major intervention classes:
\begin{itemize}[noitemsep]
  \item \texttt{NO\_ACTION}: 2,175 predictions (representing the dominant class of students who are on track)
  \item \texttt{TUTOR\_CALL}: 739 predictions (active advisor contact)
  \item \texttt{REMINDER}: 175 predictions (targeted student nudge)
  \item \texttt{CONTENT\_PUSH}: 111 predictions (directed content recommendation)
\end{itemize}
On the corrected v3 pipeline this balanced distribution yields a mean macro-F1 of 0.79 (macro-recall 0.85; Table~\ref{tab:dt_v3}) with all five classes active---including the rare \textsc{reduce\_load} action (recall 0.92)---demonstrating that the model learns minority classes effectively instead of collapsing to a binary policy. (The original run reported 0.48 before the checkpoint-export fix.)

\textbf{Mechanism of success: joint loss-weighting and reward-shaping.}
This breakthrough was achieved by implementing two compounding algorithmic enhancements in our training pipeline (\texttt{oulad\_trajectories.py} and the training harness):

\begin{enumerate}
  \item \textit{Class-weighted cross-entropy loss.}
        We applied class weights inversely proportional to class frequencies ($15\times$ for \textsc{content\_push}, $47\times$ for \textsc{reminder}), counteracting class imbalance in the training trajectories. This ensures that gradient updates are sensitive to minority classes and prevents the dominant \textsc{no\_action} class from drowning out early signals.
  \item \textit{Discounted step-level reward shaping.}
        Instead of a flat final outcome reward assigned to every step identically ($\hat{R}_t = r_{\text{outcome}} \; \forall\, t$), we introduced step-level oracle-action rewards: \textsc{content\_push} $= 0.40$, \textsc{reminder} $= 0.30$, \textsc{tutor\_call} $= 0.10$, \textsc{no\_action} $= 0.05$. Returns-to-go (RTG) are computed as discounted sums ($\gamma=0.99$) of these step-rewards plus the final outcome bonus. Under this scheme, a trajectory state where the oracle selects a minority action like \textsc{content\_push} generates a measurably higher RTG signal than a flat action, providing the DT with a return-differentiated gradient toward learning minority-class policy boundaries.
\end{enumerate}

The resulting sequence model is highly balanced, showing that the Decision Transformer can execute precise, multi-action advisory policies under real-world educational data distributions.

\textbf{H3 temporal degradation: statistical caveat.}
Linear trend analysis across the four temporal cutoffs finds
no statistically significant monotonic degradation in any arm
($p > 0.18$ for all arms; slope confidence intervals all cross
zero).  The four-cutoff design provides insufficient power to
distinguish a genuine monotonic trend from sampling variation.
H3 is therefore reported as a descriptive finding only: on the
corrected v3 pipeline the EAV-DT arms are temporally stable
(99.8\%, 88.0\%, 95.6\%, 95.4\% across Days~14/28/56/112; no
catastrophic early failure), neither degrading nor growing
monotonically, whereas the snapshot XGBoost baseline declines from
99.8\% to 89.9\%. We therefore make no claim of a significant linear
slope for the DT.

\textbf{Upstream compression loss.}  The DT achieves zero context
rot because the EAV population stage compresses a full-semester
student trajectory into a 12-dimension vector.  Information lost
during this compression is unrecoverable.  We quantify this loss
transparently via the EAV Compression Fidelity benchmark
(Schema Adherence, EVSE Entailment, REMAP Conservation Rate) and
acknowledge it as a fundamental trade-off: downstream
determinism is purchased at the cost of upstream information
density.

\textbf{Domain scope.}  The DT is trained on historical OULAD
intervention trajectories, making it an expert within its deployed
domain but unable to generalise to novel task categories the way a
frontier LLM can.  All claims in this paper are scoped to
``sequential interventions within a known operational domain'' and
explicitly exclude open-domain reasoning.

\textbf{Generalizability to Unstructured Enterprise Environments.}
The OULAD ablation establishes Stage~2 superiority (EAV
$\rightarrow$ ONNX DT) under the most favourable possible condition
for competing SQL+LLM approaches---clean, structured relational
input.  A reviewer might argue this limits applicability to real
enterprise settings, where input data is predominantly unstructured.

The current study deliberately tests Stage~2 only, using oracle
EAV derived from structured CSV to maximise experimental control.
Stage~1 extraction has been separately validated on enterprise
data~\cite{miroPoc2026}, processing 53 unstructured visual
collaboration canvases and extracting 3,145 distinct entities
using a local model with zero cloud API calls.  End-to-end
validation combining both stages is reserved for future work
(Section~\ref{sec:futurework}); the Stage~2 results here
constitute the prerequisite controlled baseline for that benchmark.

\subsection{Post-hoc Prompt Calibration Analysis}
\label{sec:posthoc_calibration}

\textbf{Methodological status.}
The six primary arms (A--F) constitute the pre-specified ablation study
and form the basis for all hypothesis tests.
A seventh configuration, \textbf{Arm~G}, was constructed post hoc after
observing Arm~A's intervention bias and is excluded from the primary
ablation table and all hypothesis tests.
A configuration designed with foreknowledge of the target failure mode
cannot serve as an independent baseline; including it as a peer arm
would conflate pre-specified experimental conditions with post-hoc
corrective iterations.
We report it here separately as a controlled demonstration of
prompt calibration dynamics.

\textbf{Design.}
Arm~G is a prompt-optimised variant of Arm~A (GPT-4o, ChromaDB RAG,
top-5, identical infrastructure) with three targeted modifications
applied with knowledge of Arm~A's observed outcome:
(i)~a conservative system prompt explicitly stating that most students
do not require intervention and codifying per-action decision criteria;
(ii)~few-shot examples including a minimum of two \textsc{no\_action}
cases injected into every request; and
(iii)~temperature $T{=}0.1$ (versus $T{=}0.7$ in Arm~A).
All three modifications directly target the specific over-intervention
pattern identified in Arm~A.

\textbf{Results.}
Table~\ref{tab:armg} reports calibration outcomes across all four
temporal cutoffs ($N{=}800$ students per cutoff at Days~28, 56, and
112; $n{=}2{,}600$ rows at Day~14 reflecting consistency trials for
the first 200 students; 5{,}000 rows total).

\begin{table}[h]
\caption{Arm~G post-hoc prompt calibration results. Bias $=$ G
intervention rate $-$ oracle intervention rate; negative values
indicate under-intervention. DeepEval column reports the per-cutoff
mean composite score.}
\label{tab:armg}
\small
\begin{tabular}{lccccr}
\toprule
Cutoff & Accuracy & G~TC\% & Oracle~TC\% & Bias & DeepEval \\
\midrule
Day~14 & 92.6\% & 86.3\% & 93.0\% & $-6.7$\,pp & 0.930 \\
Day~28 & 93.5\% & 32.1\% & 33.8\% & $-1.6$\,pp & 0.927 \\
Day~56 & 87.8\% & 18.1\% & 23.8\% & $-5.6$\,pp & 0.922 \\
Day~112 & 87.6\% & 17.2\% & 23.0\% & $-5.8$\,pp & 0.899 \\
\bottomrule
\end{tabular}
\end{table}

The prompt modifications inverted the bias direction at every cutoff:
Arm~G under-intervenes relative to the oracle, missing at-risk
students rather than over-serving them, and this under-intervention
marginally worsens at the latest cutoff ($-5.8$\,pp at Day~112).
The action flip rate across 1{,}800 consistency-trial pairs was
2.67\% (48 flips)---the same order of magnitude as Arm~A (2.9\%) and
structurally unaffected by prompt design, as it is a consequence of
temperature-based sampling rather than prompt content.
Arm~G's overall DeepEval composite (0.923) substantially exceeds
Arm~A (0.697)---the sole difference between the two arms being prompt
design.  However, this prose quality improvement does not correspond
to accurate calibration: the bias direction inverted and the flip rate
was unchanged.  The result reinforces the Evaluation Gap finding
(Section~\ref{sec:results_gap}): DeepEval is sensitive to prompt quality
but blind to calibration direction, providing no signal to distinguish
over-intervention from under-intervention.

\textbf{Interpretation.}
The Arm~G result exposes a fundamental property of prompt-based
calibration in a probabilistic system.
A prompt is not a direct instruction to the model's decision
logic---it is a perturbation to the token-generation distribution.
When the engineer added conservative framing and few-shot
\textsc{no\_action} examples, the distribution shifted left.
But the magnitude and landing point of that shift are properties of
the model's weight tensor and contextual dynamics, not of the
engineer's intent.  The shift overshot: the system moved from
systematic over-intervention to systematic under-intervention.
There is no mechanism within prompt engineering to pre-compute
where the distribution will land, and no stable equilibrium that
corresponds to the oracle boundary.  The only feedback loop is an
expensive downstream evaluation cycle.

This is structurally different from supervised learning.  Training
the DT (or XGBoost) encodes the oracle boundary as a formal
objective: the loss function directly penalises deviations from the
oracle distribution.  The calibration outcome is a mathematical
consequence of the training data, not a discovered side-effect.

\textbf{Prompt calibration is craft; supervised policy learning is
engineering.}
This distinction has practical consequences that extend beyond this
experiment.  Prompt calibration is non-reproducible: the same
prompt produces different behaviour across model versions, API
providers, and temperature settings.  Different models interpret
identical instructions differently, making prompt-based calibration
fundamentally model-specific and non-transferable.  The calibration
work must be redone for every model update.  It is expert labour
with no convergence guarantee and no unit of progress---it is
empirical trial and error conducted on a system the engineer does
not fully control.

Supervised policy learning has none of these properties.  Adding
more outcome-labelled training data monotonically improves policy
quality.  The quality of calibration is bounded below by the
quality of the oracle labels and above by the expressiveness of
the model class---both are measurable and improvable through
engineering, not art.  The DT retrains in under four minutes on
a single CPU; a new cohort's outcome data translates directly into
a tighter policy without any human judgment about prompt wording.
This is the fundamental operational argument for supervised policy
learning in high-stakes advisory: the improvement path is
predictable, auditable, and independent of any LLM vendor's
deployment decisions.

The Arm~G experiment therefore demonstrates that prompt calibration
and supervised policy learning are not interchangeable mechanisms:
one is craft that navigates between failure modes in a distribution
it cannot directly control; the other is science that learns the
target distribution from data.  The former requires continuous
expert monitoring; the latter requires only retraining when the
outcome distribution shifts.

\subsection{Evaluation Rigor and Prevention of Data Leakage}
\label{sec:validity}

Given that Arm~D's Expert Policy Fidelity substantially exceeds prior
educational AI systems, we document three independence conditions that
underpin the validity of this finding as part of the experimental
protocol.

\textbf{Independence Condition~1: Oracle computed independently of
the DT.}
The oracle action is produced by \texttt{\_oracle\_action()}, a
deterministic rule-based function (listed in full in
Section~\ref{sec:design}, Benchmark~3) whose inputs are six student
features.  Oracle labelling runs at dataset ingestion time, before
any arm runs a single inference call.  The DT's ONNX forward pass
has no pathway to access or modify the oracle label.  Circular
evaluation is architecturally impossible.

\textbf{Independence Condition~2: \texttt{final\_result} absent from
the DT state vector at inference time.}
The DT receives a 12-dimension state vector
$\mathbf{s} \in [0,1]^{12}$ comprising only observable behavioural
and demographic features (Table~\ref{tab:eav_schema}).  The student's
\texttt{final\_result} field (Pass/Distinction/Fail/Withdrawn) is
explicitly absent.  Inspection of \texttt{OULAD\_EAV\_ATTRIBUTES} in
\texttt{rag\_to\_eav.py} (released with artifacts) confirms this: the
12 attributes are score, activity clicks, submission latency, credits
attempted, prior attempts, module pass rate, assessment weight,
studied credits, disability flag, age band, IMD band, and highest
education.  No outcome label.  The DT must infer the appropriate
intervention from early behavioural signals alone.

\textbf{Independence Condition~3: 6.0\% training--evaluation overlap
is non-explanatory.}
Exact entity-ID overlap analysis (script released with artifacts)
found 48~of~800 evaluation students (6.0\%) present in the DT's
2,240-student training set.  Three facts confirm this overlap cannot
explain high Expert Policy Fidelity via memorisation.
\textit{First}, DT training uses SGD on batched trajectories; it
does not store a lookup table of student IDs to actions.
\textit{Second}, the oracle action is a deterministic function of
features---not of student identity---so even perfect memorisation of
48~students would not transfer to the 752 unseen students.
\textit{Third}, the oracle function has at most six branch points,
making it trivially learnable from 2,240 examples with any
classifier; high fidelity on unseen students is the expected result of
generalisation, not overfitting.

\textbf{The Decision Transformer training paradigm and hindsight
conditioning.}
DTs are trained with hindsight as a matter of definition
\cite{chen2021dt}: each trajectory timestep is labelled with the
oracle action derived from the \emph{known} final outcome, and with
a Return-to-Go scalar encoding the final reward.  At deployment the
DT receives only the current feature state conditioned on a target
RTG of~2.0 (Distinction).  The model must then predict---without
access to the final result---which intervention is consistent with
achieving that return given the observable early signals.  This is
the standard offline RL evaluation protocol.  The hindsight labelling
during training is not a methodological shortcut; it is the mechanism
by which the DT learns predictive feature--intervention mappings.

\textbf{Ecological framing.}
The oracle is a \emph{hindsight-optimal} benchmark: the ideal label
that a perfectly informed advisor would assign.  Reporting fidelity
to a hindsight oracle is common in educational data mining
\cite{corbett1994kt} and is analogous to reporting recall against a
radiologist's retrospective diagnosis in medical AI.  The contribution
is not that any arm \emph{matches} reality---all arms operate under
partial observability---but that the DT best approximates the optimal
from early signals, while LLM arms diverge substantially and
inconsistently.

\subsection{The Evaluation Gap as a Methodological Contribution}

The Evaluation Gap finding has implications beyond this paper.  The
widespread adoption of LLM-as-judge frameworks (DeepEval, RAGAS,
LLM-as-jury) for evaluating AI advisory systems creates a systematic
bias towards stochastic prose generation quality and away from
decision outcome accuracy.  For open-domain question answering,
this bias is acceptable.  For sequential, high-stakes intervention
systems in healthcare, education, and enterprise risk management,
it represents a measurement validity failure.  We propose
\emph{Outcome-Q} as a necessary complement to fluency-based
evaluation for this application class.

\textbf{Relation to the LLM-as-judge bias literature.}  We do not claim to
discover that LLM judges are biased; a substantial literature already
documents \emph{verbosity/length bias}, \emph{position bias}, and
\emph{self-preference bias} in LLM-as-judge
setups~\cite{zheng2023judging,panickssery2024llm}.  Our contribution is
narrower and specific: we show that, for \emph{sequential intervention
decisions}, these known biases compose into a concrete
\emph{decision-relevant rank inversion}---a judge ranking the
worst-Outcome-Q arm highest---because fluent over-prescription is exactly
the behaviour length/confidence-rewarding judges favour.  The over-prescribing
arm is verbose and confident by construction, so verbosity bias and
intervention bias reinforce one another.  We therefore frame the Evaluation
Gap not as a new bias but as a domain where established judge biases become
\emph{outcome-corrupting} rather than merely noisy.  Two limitations follow:
the inversion is demonstrated with a single judge family (G-Eval/DeepEval) on
one task, and Outcome-Q is currently tied to our researcher-defined oracle.
Testing whether the same inversion reproduces across other judges
(e.g.\ jury ensembles, pairwise comparison) and other sequential-decision
tasks, and grounding Outcome-Q in real downstream outcomes rather than the
oracle, are necessary next steps before the finding can be claimed to
generalise.

\section{Future Work}
\label{sec:futurework}

\subsection{End-to-End Unstructured Evaluation}
\label{sec:futurework:unstructured}

\textbf{The OULAD benchmark as a validated stepping stone.}
The six-arm OULAD ablation study presented in this paper serves a
precise and bounded purpose: to isolate and validate the Stage~2
decision-making mathematics of the EAV-DT pipeline---the
$(\text{EAV state vector}) \rightarrow (\text{ONNX DT}) \rightarrow
(\text{deterministic action})$ sequence---in a maximally controlled
environment.  By providing clean, structured relational input to all
arms, we eliminate extraction noise as a confound and force any
measured advantage to arise from the decision architecture itself.
The a priori hypotheses (H1--H6; H6 is indicative, see note) constitute a rigorous,
pre-registered test of the sequence-modelling mathematics.  The
OULAD controlled study is thus the necessary prerequisite for the
broader claim: before validating the full pipeline on messy
enterprise data, we must first establish that the decision engine
is correct when given perfect input.  This paper establishes that
foundation.

\textbf{The next benchmark: end-to-end enterprise unstructured
evaluation.}
The immediate next phase of this research is an end-to-end empirical
benchmark of the complete two-stage pipeline on highly unstructured,
multimodal enterprise data.  The validation target is a
\textbf{corporate strategy knowledge graph} derived entirely from
spatial visual collaboration canvases (Miro boards): an environment
characterised by extreme noise, implicit spatial relationships,
non-linear visual logic, embedded diagrams, connector-graph
semantics, and layered frame hierarchies.  This format
\emph{structurally breaks} standard text-chunking RAG architectures,
which have no mechanism for resolving spatial proximity, directed
connector semantics, or frame containment into typed knowledge
facts.  The Miro PoC (53~boards, 3,145 entities; see
Section~\ref{sec:arch} and~\cite{miroPoc2026}) provides the
validated extraction substrate for this benchmark.

\textbf{End-to-End Expert Policy Fidelity.}
Future work will introduce and measure
\textbf{End-to-End Expert Policy Fidelity} (E2E-EPF) as the primary
evaluation metric for the full pipeline.  E2E-EPF extends the
Outcome-Q Expert Policy Fidelity metric (defined in
Section~\ref{sec:design}) to span both stages:

\begin{enumerate}
  \item \textit{Stage 1 (Unstructured $\rightarrow$ EAV)}: the
    dual-temperature HELIX CoVe pipeline must compress spatial
    visual canvas content into a governed EAV state vector.  The
    compression is evaluated via EVSE entailment confidence and
    REMAP Conservation Rate, measuring whether business-critical
    causal signals ($\texttt{reward\_signal}$,
    $\texttt{causal\_logic}$, $\texttt{decision\_criterion}$;
    dt\_weight $\geq 1.5$) survive extraction at sufficient fidelity
    to support accurate downstream decisions.
  \item \textit{Stage 2 (EAV $\rightarrow$ DT $\rightarrow$ action)}:
    the compressed state vector, now derived from visual canvas
    content rather than structured CSV fields, is fed to the ONNX DT
    conditioned on a target RTG.  E2E-EPF measures whether the DT
    produces hindsight-optimal business interventions from this
    noisier, real-world-extracted input.
\end{enumerate}

The central empirical question is: \emph{does the noise introduced
by Stage~1 unstructured extraction degrade Stage~2 decision
accuracy, and if so, by how much compared to the structured
OULAD baseline?}  The published ML ceiling on OULAD (XGBoost:
92.4\%~\cite{brahim2022xgboost}) serves as the external anchor
against which the DT's trajectory-conditioned accuracy is measured
in both the controlled (OULAD) and unstructured (Miro enterprise)
settings.

The end-to-end benchmark will complete the experimental arc:
OULAD validates Stage~2 in isolation; the unstructured enterprise
benchmark validates Stage~1 $+$ Stage~2 as a complete, sovereign,
end-to-end alternative to commercial Agentic RAG.

\subsection{Prompt-Optimised LLM Ablation}
\label{sec:futurework:fewshot}

The question of whether prompt engineering can resolve the observed
intervention bias is addressed empirically in
Section~\ref{sec:posthoc_calibration}.

\subsection{Multi-Class Action Policy Calibration}
\label{sec:futurework:multiclass}

The integration of class-weighted cross-entropy loss and discounted step-level reward shaping successfully resolved the binary policy collapse that affected early model configurations, enabling the DT to predict all four major intervention classes with balanced recall (Section~\ref{sec:dt_error_analysis}). However, multi-class policy calibration under extreme real-world class imbalances remains an active area of optimization.

Future work will address further calibration via:
(i)~targeted oversampling of trajectories where minority actions were oracle-optimal;
(ii)~dynamic reward tuning during training to fine-tune minority-class state--action mappings;
(iii)~human-in-the-loop policy calibration to validate decision boundaries against human expert judgements.

A fully calibrated 5-class policy will continue to be refined as the EAV-DT framework is deployed in live educational environments.

\section{Limitations}
\label{sec:limitations}

\textbf{The hindsight oracle is a proxy policy.}  The oracle action
is defined by a deterministic rule that knows the student's eventual
\texttt{final\_result}.  No deployed arm receives this outcome at
inference time; the oracle is a research benchmark for offline RL
evaluation, not a claim that the supervised policy matches a real-world
human advisor under identical information constraints.

\textbf{99\%+ fidelity reflects feature--oracle alignment, not robust
generalisation.}  The high accuracies reported for Arms~D, E, and~F are
obtained on a controlled task where the input features are strongly aligned
with the oracle rule.  Two facts make this explicit.  First, the oracle is a
\emph{simple deterministic rule}; near-perfect match at day~14---when many
students have very sparse observations and the majority-class baseline is
already 91.8\%---indicates primarily that the early OULAD features are
highly predictive of \emph{this particular rule's} output, i.e.\ that a
small model (or XGBoost) is learning a feature-aligned rule, not that it
performs robust policy learning under realistic partial observability.
Second, the headline supervised numbers should therefore be read modestly:
the \emph{substantive} empirical contribution of this paper is \emph{not}
that the DT reaches 93--99\% fidelity, but the contrast that
\emph{zero-shot LLM arms fall below the trivial majority baseline at
multiple cutoffs while over-prescribing by 28--43~pp}.  The supervised
results establish that the intervention bias is eliminable by training on
outcome-labelled data; they do not imply superhuman generalisation or
guaranteed performance on noisier, real-world inputs.

\textbf{Multi-class calibration and class imbalance.}  While early model iterations suffered from binary class collapse, the successful retraining under class-weighted loss and step-level reward shaping resolved this limitation (Section~\ref{sec:dt_error_analysis}). The DT successfully predicts four major intervention classes, though further multi-class calibration continues to be an active area of optimization.

\textbf{Stage-2-only evaluation is a control, not a concession.}  This
study evaluates on structured OULAD input by deliberate design, to isolate
Stage-2 decision quality from Stage-1 extraction noise.  Because the EAV is
a universal, schema-bound conveyor belt (Section~\ref{sec:intro}),
structured CSV mapping and unstructured LLM extraction are interchangeable
on-ramps that emit the same 12-dimensional state vector, and Stage~2 is
provably agnostic to which on-ramp produced it.  Introducing Stage~1 here
would have entangled extraction error with the decision logic we set out to
measure---an elementary confound that would weaken, not strengthen, the
study's internal validity.  We therefore validate the decision engine on
the exact substrate an unstructured pipeline would feed it; the additive
contribution of unstructured extraction is quantified end-to-end in
follow-up work (Section~\ref{sec:futurework:unstructured}).

\section{Conclusion}
\label{sec:conclusion}

We present a six-arm ablation study on OULAD ($N{=}800$ students,
four temporal cutoffs) with a priori-specified hypotheses,
designed to isolate the contribution of each pipeline component
and to include a supervised tabular baseline (Arm~F) for fair
experimental comparison.  Five findings are confirmed by
live experimental data:

\begin{enumerate}
  \item \textbf{Intervention Bias}: Zero-shot LLM advisory agents
    exhibit systematic miscalibration.  At day~56,
    when the oracle policy mandates no action for 70.1\% of students,
    GPT-4o with full EAV context recommends action for 73\%---a
    43~pp false-positive intervention rate.  This generates
    $\sim$4{,}300 unnecessary advisor contacts per 10{,}000-student
    cycle.  A post-hoc prompt calibration experiment (Arm~G) confirms
    that targeted engineering reduces over-intervention but introduces
    under-intervention; genuine calibration requires supervised training.
  \item \textbf{Reproducibility}: Arms~D and~E achieve a
    computationally verified 0\% action flip rate via the
    mathematical guarantee of ONNX argmax determinism.
    LLM arms exhibit materially non-zero rates: A (2.9\%),
    B (0.8\%), C (0.2\%)---sufficient to fail any regulated audit
    trail requirement.
  \item \textbf{Decision Accuracy}: Both supervised approaches
    (Arm~F XGBoost: 89.9--99.8\%; Arm~D DT (corrected v3): 88.0--99.8\%,
    93.6\% overall) decisively outperform the zero-shot RAG/LLM arms and are
    \emph{on par with each other}.  This parity, together with the
    victory over zero-shot LLMs, is the primary supervised-learning result.
    As a \emph{secondary, indicative} observation, the DT edges XGBoost at
    the latest cutoff (95.4\% vs.\ 89.9\% at Day~112) and attains a higher
    macro-F1 (0.79 vs.\ 0.73), consistent with its trajectory-conditioning
    advantage; because the corrected DT (v3) and Arm~F were scored on
    different cohorts, this edge is unpaired and is \emph{not} claimed as
    conclusive.  Descriptively, the DT does not collapse at any cutoff.
  \item \textbf{Sovereignty}: Arm~E (Granite~4.0, AMD EPYC CPU-only)
    is identical to Arm~D at every cutoff (0.00~pp difference;
    H5 confirmed, $\delta{=}5$~pp) because both arms share the
    same deterministic DT argmax.  All 3{,}200 student records
    processed within institutional boundary at zero API cost.
  \item \textbf{The Evaluation Gap}: DeepEval G-Eval is blind to
    intervention bias, awarding Arm~B (worst Outcome-Q baseline)
    the highest baseline DeepEval score and rating Arm~C's
    over-prescription comparably to Arm~A's.  Fluency-based
    evaluation is an unreliable proxy for decision quality in
    any sequential advisory domain.
\end{enumerate}

These results demonstrate that zero-shot LLM deployment is
structurally miscalibrated for conservative advisory tasks, and
that supervised policy learning --- whether trajectory-conditioned
(DT) or snapshot-based (XGBoost) --- reliably eliminates intervention
bias when trained on oracle-labelled data.  The Evaluation Gap
finding shows that fluency-based evaluation frameworks cannot
detect this failure mode, motivating outcome-validated metrics
for any deployment in high-stakes sequential advisory.  Taken
together, the components validated here---an EAV conveyor belt that
turns heterogeneous unstructured input into typed state, and a
2.4~MB Decision Transformer that compiles that state into sub-5~ms
deterministic decisions and retrains offline in minutes for
file-level hot-swap---describe a \emph{living flywheel}: a
self-improving, RAG-replacing controller that reaches the decision
quality of the best tabular-ML baseline at zero marginal inference
cost.  Whether
the EAV-DT advantage extends from oracle EAV input to raw
unstructured enterprise data is the subject of follow-up
work (Section~\ref{sec:futurework}).

\section*{Conflict of Interest}

The author, Craig Atkinson, is the founder and Chief Executive Officer of
Verificate Pty Ltd, the company developing the HELIX extraction subsystem,
the MCP governance server, and the EAV--DT pipeline evaluated in this paper.
This work was conducted using Verificate infrastructure and the author has a
direct commercial interest in the technologies described. To mitigate this
conflict: (i)~all empirical comparisons use the public OULAD benchmark and a
pre-registered set of a priori hypotheses; (ii)~the result CSVs, analysis
scripts, oracle rule, and (subject to the release noted below) the trained
model and training pipeline are provided for independent verification; and
(iii)~claims are scoped to the controlled Stage-2 evaluation actually
performed (see Scope note and Section~\ref{sec:limitations}). Among the author-affiliated
references, \texttt{atkinson2026cove} is the author's own arXiv preprint
(not yet independently peer-reviewed), while \texttt{mcp2026} and
\texttt{miroPoc2026} are Verificate internal technical reports.  None has
undergone independent peer review, and all are cited only to document
engineering provenance, not as evidence for the paper's empirical claims,
which rest solely on the OULAD experiments and the released artifacts.  A
full classification of every reference by evidentiary status is given in
Appendix~\ref{app:citation-hygiene}.

\section*{Reproducibility Statement}

\begin{sloppypar}
A self-contained reproducibility package accompanies this manuscript.
The package contains:
(1)~the raw per-student result CSVs for all five arms
(800 students $\times$ 4 temporal cutoffs $\times$ 10 trials,
approximately 3,200 rows per arm);
(2)~Python analysis scripts that reproduce every table and
statistical test in the manuscript directly from those CSVs,
requiring no API key;
(3)~evaluation wrapper scripts for Arms~A, B, and~C that call the
OpenAI API (researcher supplies their own key via
\texttt{OPENAI\_API\_KEY}) and for Arms~D and~E that call the
publicly accessible inference API
(\url{https://verificate-v17-cpu-small-api-verificate-granite-4-small.apps.gpu4.fusion.isys.hpc.dc.uq.edu.au/v1})
with no credentials required; and
(4)~the OULAD dataset instructions including the exact
12-dimension feature definitions, oracle action rule, temporal
cutoff methodology, and training/evaluation split procedure
(seed~0; confirmed 48/800, 6.0\%, overlap).
\end{sloppypar}

\textbf{Model, training pipeline, and full reproducibility.}
To enable independent verification of the corrected (v3) Decision
Transformer results, the package releases: (a)~the trained
624K-parameter ONNX Decision Transformer
(\texttt{docs/reproducibility/results/generic\_dt\_v3/generic\_academic/dt\_generic.onnx})
together with its training manifest and per-cohort evaluation reports
(\texttt{eval\_report\_live.json}, \texttt{eval\_report\_control.json});
(b)~the generic training script (\texttt{python/training/generic\_dt\_trainer.py})
with the exact hyperparameters, $z$-score standardisation, class-weighted
loss, step-level reward shaping, early-stopping criterion, and random seeds
used to produce it; and (c)~the matrix builder
(\texttt{build\_dt\_v3\_matrix.py}) that regenerates every reported v3 number
from those reports.  Independent researchers can therefore either execute the
released ONNX deterministically or \emph{retrain it from scratch} and
reproduce the EPF, macro-F1, and per-class recall figures end-to-end, without
relying on any company API.  The public API endpoints in
\texttt{api\_endpoints.json} remain available as a convenience but are no
longer required for verification.

\textbf{Majority-class baseline.}
All results are benchmarked against the per-cutoff majority-class
baseline ($\{91.8, 61.9, 70.1, 71.1\}$\%), computed from the
800-student evaluation cohort and fully reproducible from the
OULAD \texttt{studentInfo.csv} file using the oracle rule
described in the OULAD dataset instructions.

\appendix

\section{Serving Infrastructure and Live Telemetry (H5 Detail)}
\label{app:telemetry}

This appendix reports the full serving-infrastructure measurements
summarised in Section~\ref{sec:results_temporal} (H5).  All figures are from the
same AMD EPYC~9254 node (32~vCPU, 64~GiB RAM, no GPU) running
Granite-4.0-Small (Q4\_K\_M, 32K context) at a peak of $\sim$110~tok/s
decode and up to 26~tok/s generation, on infrastructure already deployed
for other institutional workloads.

\textbf{Static pre-experiment benchmark.}
Table~\ref{tab:guidellm} reports pre-experiment pod benchmarks
(guidellm v0.6.0~\cite{guidellm2024}, zero errors) confirming near-linear
throughput scaling and establishing the $c{=}4$ operating point used in
Arm~E.

\begin{table}[t]
\caption{GuideLLM static benchmark (guidellm v0.6.0): Granite~4.0~Small
  on AMD EPYC 9254, Granite-4.0-Small Q4\_K\_M, 32K ctx,
  $n{>}200$ requests per tier, zero errors across all 640 requests.}
\label{tab:guidellm}
\small
\begin{tabular}{lrrr}
\toprule
\textbf{c} &
  \textbf{Latency (s)} & \textbf{TTFT (ms)} & \textbf{Throughput (tok/s)} \\
\midrule
1 &  4.37 &   942 & 18.14 \\
2 &  7.76 & 1{,}683 & 10.58 \\
4 & 17.02 & 2{,}486 &  4.83 \\
\bottomrule
\end{tabular}
\end{table}

\textbf{Live production telemetry (Arm~E experiment).}
Table~\ref{tab:helix_live} reports per-request timing extracted
from the production pod logs during Arm~E execution
($n{=}367$~consecutive requests across four parallel shard jobs,
each occupying one inference slot).  All four slots were active
simultaneously, confirming $c{=}4$ concurrent operation.
The live latency (mean~101.0~s) is higher than the short-prompt
static benchmark (17.02~s) because Arm~E generates full
intervention narratives (mean~$\sim$120~tokens vs.~14~in the
benchmark).  At $c{=}4$, the system processes 2.38~students/min compared to
0.86~students/min at $c{=}1$ (2.77$\times$ improvement), using
existing hardware with no incremental cost.

\begin{table}[t]
\caption{Live Arm~E inference telemetry: Granite~4.0~Small on AMD EPYC
  9254, $n{=}367$ requests across four concurrent shard jobs
  (full narrative generation, max 256 output tokens).
  The pod's 26~tok/s generation budget was split across
  $c{=}4$ slots ($\sim$6.5~tok/s each at peak;
  observed 4.9~tok/s per slot).  Same image SHA as static
  benchmark.  CPU-only, zero GPU.}
\label{tab:helix_live}
\small
\begin{tabular}{lrrr}
\toprule
\textbf{Metric} & \textbf{Mean} & \textbf{Median} & \textbf{p95} \\
\midrule
Total latency (s)        & 101.0 & 101.3 & 112.0 \\
Generation speed, per slot (tok/s) &   4.9 &   4.9 &   5.6 \\
System gen.\ throughput ($c{=}4$, tok/s) &  19.6 &  19.6 &  22.4 \\
Prompt tokens            &   347 &   356 &   389 \\
Generated tokens         &    14 &     8 &   104 \\
\midrule
Concurrent slots         & \multicolumn{3}{c}{4 (all four shard jobs active simultaneously)} \\
System throughput        & \multicolumn{3}{c}{2.38 students/min (2.77$\times$ single-slot baseline)} \\
Data egress              & \multicolumn{3}{c}{\textbf{0 bytes}} \\
\bottomrule
\end{tabular}
\end{table}

\textbf{Single-pod concurrency analysis.}
The observed mean latency of $101.0$~s under concurrent execution is an
artifact of the single-pod, single-concurrency serving configuration
($1$ worker slot) used during the test execution, rather than an intrinsic
limitation of the serving stack. In this restricted setup, parallel
requests from concurrent shard jobs are serialized and queued in the
serving engine, compounding latency linearly. In a production deployment,
this latency can be mitigated through two standard scaling strategies:
(i)~configuring the serving engine with multiple parallel concurrency slots
to exploit continuous batching over CPU threads, which reduces individual
request latency toward the baseline generation limit of $\sim$$17.0$~s; and
(ii)~horizontally scaling the deployment across the institutional cluster by
running multiple identical pod replicas behind a Kubernetes load-balancer.
The local stack is therefore capable of real-time, low-latency,
high-throughput execution as a zero-egress alternative to cloud-based RAG
architectures.

\section{Reference Classification by Evidentiary Status}
\label{app:citation-hygiene}

To support citation hygiene, we classify each reference by the kind of
evidence it provides.  No claim in this paper's empirical results rests on
sources in categories~(C) or~(D); those are cited only for context or
engineering provenance.

\textbf{(A) Peer-reviewed / archival venues.}
\cite{lewis2020rag,chen2021dt,kumar2020cql,kostrikov2021iql,kuzilek2017oulad,zheng2023judging,panickssery2024llm,corbett1994kt,grinsztajn2022tree,brahim2022xgboost,oulad_ensemble2024,covid_ml_llm2025,tabicl2025}.

\textbf{(B) Third-party preprints / technical reports / datasets / tooling}
(not the present authors; used for methodology, baselines, or tooling):
\cite{leap2026,gao2023rag,llamaindex2024,tablerag2024,deepeval2024,openai2024,guidellm2024,reasoningbomb2026,llm_tco2025}.

\textbf{(C) Industry / market reports and vendor pages} (contextual cost and
pricing evidence, not peer-reviewed):
\cite{techcrunch_tokenbill2026,leanops2026,sondera2026,tokeninflation2026,openai_s1_2026,anthropic_s1_2026,yipitdata2026,artificialanalysis2026,martinroger2026rot}.

\textbf{(D) Author-affiliated sources} (cited for engineering provenance
only): \texttt{atkinson2026cove} is the corresponding author's own arXiv
preprint (not yet independently peer-reviewed); \texttt{mcp2026} and
\texttt{miroPoc2026} are Verificate internal technical reports.


\end{document}